\documentclass[11pt]{article}

\usepackage[final]{acl}

\usepackage{times}
\usepackage{latexsym}

\usepackage[T1]{fontenc}

\usepackage[utf8]{inputenc}

\usepackage{microtype}

\usepackage{inconsolata}

\usepackage{graphicx}
\usepackage{tcolorbox}
\usepackage{enumitem}
\usepackage{geometry}
\usepackage{amssymb}
\usepackage{mathrsfs}
\usepackage{amsmath}
\usepackage{hhline}
\usepackage{amsfonts}
\usepackage{multirow}
\usepackage{tabularx}
\usepackage{array}
\usepackage{booktabs}
\usepackage{verbdef}
\usepackage{colortbl}
\usepackage{color}
\usepackage{subcaption}
\usepackage{threeparttable}
\usepackage{float}
\usepackage{caption}
\usepackage{arydshln}
\usepackage{rotating}
\usepackage{xcolor}
\usepackage{placeins}
\usepackage{makecell}
\usepackage{afterpage}
\usepackage{blindtext}
\usepackage{pifont}
\usepackage{threeparttablex}
\usepackage[framemethod=TikZ]{mdframed}
\usepackage{hyperref}
\usepackage{marvosym}

\title{Disco-RAG: Discourse-Aware Retrieval-Augmented Generation}

\author{
 \textbf{Dongqi Liu\textsuperscript{$\Omega$}\textsuperscript{$\Theta$}\thanks{\scalebox{1.3}{\Letter} dongqi@lst.uni-saarland.de}},
 \textbf{Hang Ding\textsuperscript{$\Delta$}},
 \textbf{Qiming Feng\textsuperscript{$\Gamma$}},
 \textbf{Xurong Xie\textsuperscript{$\Psi$}},
 \textbf{Zhucun Xue\textsuperscript{$\Psi$}},
 \\
 \textbf{Chengjie Wang\textsuperscript{$\Theta$}},
\textbf{Jian Li\textsuperscript{$\Theta$}\thanks{Corresponding Author}},
 \textbf{Jiangning Zhang\textsuperscript{$\Theta$}\footnotemark[2]},
 \textbf{Yabiao Wang\textsuperscript{$\Psi$$\Theta$}\footnotemark[2]}
 \\
 \textsuperscript{$\Omega$}Saarland University,
 \textsuperscript{$\Delta$}Shanghai Jiaotong University \\
 \textsuperscript{$\Gamma$}Fudan University,
 \textsuperscript{$\Psi$}Zhejiang University,
 \textsuperscript{$\Theta$}Tencent YouTu Lab \\
}

\begin{document}

\maketitle

\begin{abstract}
Retrieval-Augmented Generation (RAG) has emerged as an important means of enhancing the performance of large language models (LLMs) in knowledge-intensive tasks. However, most existing RAG strategies treat retrieved passages in a flat and unstructured way, which prevents the model from capturing structural cues and constrains its ability to synthesize knowledge from dispersed evidence across documents. To overcome these limitations, we propose \texttt{Disco-RAG}, a discourse-aware framework that explicitly injects discourse signals into the generation process. Our method constructs intra-chunk discourse trees to capture local hierarchies and builds inter-chunk rhetorical graphs to model cross-passage coherence. These structures are jointly integrated into a planning blueprint that conditions the generation. Experiments on question answering and long-document summarization benchmarks show the efficacy of our approach. \texttt{Disco-RAG} achieves state-of-the-art results on the benchmarks without fine-tuning. These findings underscore the important role of discourse structure in advancing RAG systems. The project information is available at \url{https://dongqi.me/projects/Disco-RAG}.
\end{abstract}

\section{Introduction}

The advent of large language models (LLMs; \citealt{touvron2023llama, yang2025qwen3, achiam2023gpt}) has advanced research progress in natural language processing (NLP), achieving competitive performance across a wide range of tasks, including question answering \citep{wu2025mmqa, lee-etal-2025-rescore, zhang-etal-2025-belle}, document summarization \citep{mondshine-etal-2025-beyond-n, liu-etal-2025-talk, wang-etal-2025-empirical, luo-etal-2025-dtcrs}, and text generation \citep{duong2025scope, bigelow2025forking, que-rong-2025-pic, zhang-etal-2025-personalized}. However, due to the reliance on static training corpora, LLMs can be inadequate for knowledge-intensive scenarios, such as handling domain-specific knowledge, proprietary data, or information requiring real-time updates \citep{chang-etal-2025-main, lee-etal-2025-hybgrag, yue2025inference, wang-etal-2024-searching, xia2025mmedrag,ding-etal-2026-scirag,chen2026grorag}. Retrieval-Augmented Generation (RAG) has been proposed as a suitable strategy by integrating an external knowledge component through retrieval-based mechanisms \citep{lewis2020retrieval, asai2024selfrag, chan2024rqrag}. 

\begin{figure}[t]
  \centering
  \includegraphics[width=\linewidth]{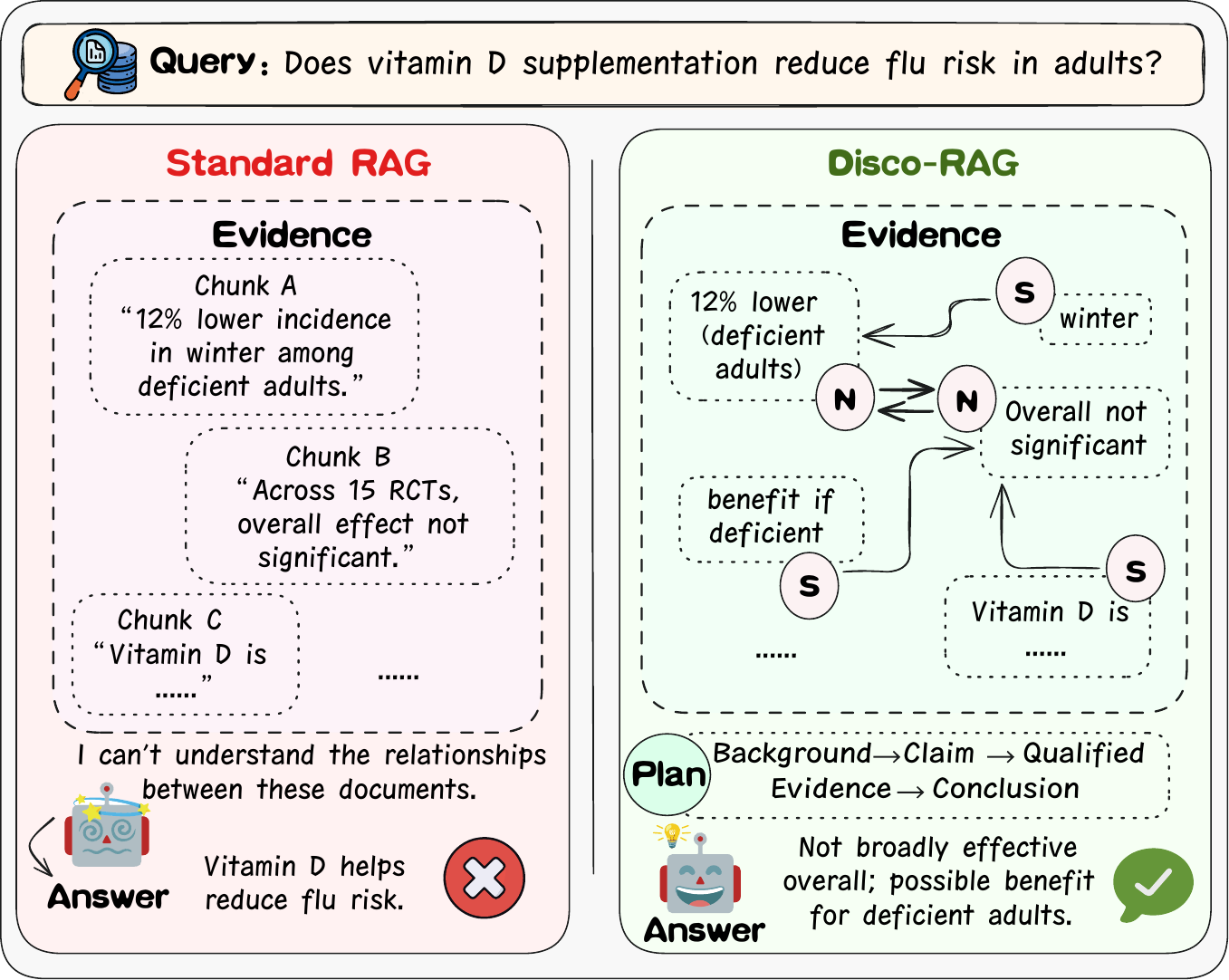}
  \caption{Comparison between standard RAG and \texttt{Disco-RAG}. While standard RAG retrieves isolated chunks without structural links, \texttt{Disco-RAG} organizes evidence into discourse structures (trees \& graphs). Here, \texttt{S} denotes \textit{Satellite} (the supplementary part), and \texttt{N} denotes \textit{Nucleus} (the core part).}
  \label{fig:rag_comparison}
  \vspace{-10pt}
\end{figure}

In standard RAG pipelines, external documents are segmented into chunks, which are then encoded into vectors and stored in a database. At query time, relevant chunks are retrieved to provide contextual grounding for the LLM \citep{lewis2020retrieval}. One important but insufficiently addressed limitation of existing RAG systems concerns \textbf{the mismatch between retrieval granularity and generative understanding}. While retrieval modules return relevant chunks, these chunks are often fragmented in discourse, resembling scattered pieces of evidence without clear logical connections \citep{edge2024local, su2025parametric}. This manifests at two levels. First, \textit{intra-chunk structural blindness}: within each chunk, RAG systems may fail to capture internal discourse. Second, \textit{inter-chunk coherence gaps}: across multiple chunks, RAG systems may struggle to identify rhetorical connections. As depicted in \autoref{fig:rag_comparison} (left), \textit{Chunk A} mentions \textit{a 12\% lower incidence}, while \textit{Chunk B} notes \textit{no significant overall effect}. Without recognizing that the former is a conditional finding (\textit{e.g.,} among deficient adults in winter), standard RAG tends to overgeneralize and incorrectly conclude that \textit{vitamin D reduces flu risk}. These deficiencies prevent effective resolution of conflicting claims, as standard RAG approaches lack the capacity to organize retrieved evidence through a higher-level causal flow. This leaves the final LLM generator to grapple with a \textit{bag of facts} rather than a coherent \textit{line of reasoning}.

Recent investigations have revealed that integrating discourse knowledge into LLMs can improve downstream performance \citep{gautam-etal-2024-discourse, pu-demberg-2024-rst} and alleviate hallucinations \citep{10.1162/TACL.a.30}. These findings highlight the drawback of relying solely on flat sequential representations and underline the benefits of discourse for context engineering \citep{ma-etal-2025-pragmatics, mei2025survey}. Building on these insights, the present work aims to investigate whether explicitly modeling and providing discourse information to the LLM can improve generation quality in the context of RAG. To answer this, we propose \texttt{Disco-RAG}, a framework that constructs local discourse trees for each retrieved chunk and infers inter-chunk coherence relations across chunks to form a rhetorical graph. To synthesize information, rather than merely concatenating it, the text generator needs not only to understand the relations between evidence but also to strategize how to present them. This requires a high-level plan to orchestrate the narrative flow. We thus introduce a discourse-aware planning module that enables the model to dynamically generate a plan to guide the generation. As shown in \autoref{fig:rag_comparison} (right), the discourse-aware process enables the model to infer that \textit{vitamin D is not broadly effective but may benefit deficient adults under specific conditions}, producing more faithful answers and aligning with the underlying evidence.

In our experiments, we evaluate \texttt{Disco-RAG} on three benchmarks, \texttt{Loong} \citep{wang-etal-2024-leave}, \texttt{ASQA} \citep{stelmakh-etal-2022-asqa}, and \texttt{SciNews} \citep{pu-etal-2024-scinews}. Consistent improvements are observed compared with standard RAG systems and state-of-the-art (SOTA) methods. On the Loong benchmark, our approach delivers an overall gain of 12.74 points in LLM Score. On the ASQA dataset, our method exceeds the best existing systems on Exact Match and ROUGE-L Score by clear margins. On the SciNews benchmark, \texttt{Disco-RAG} establishes new SOTA performance across most evaluation metrics.

\textbf{In summary, our contributions are as follows:}
\begin{itemize}[leftmargin=8pt,itemsep=1pt,topsep=1pt,parsep=1pt]
\item We present \texttt{Disco-RAG}, an inference-time strategy that explicitly injects discourse knowledge into the RAG pipeline to alleviate the discrepancy between chunk-level evidence and discourse-level reasoning.
\item We propose a modeling method that combines intra-chunk discourse trees, inter-chunk rhetorical graphs, and discourse-driven plans to capture local hierarchies, cross-passage coherence, and argumentative flow.
\item We conduct experiments on knowledge-intensive QA and summarization tasks, demonstrating consistent gains over strong RAG baselines. Analysis studies further confirm the efficacy of discourse-aware guidance in enhancing generation correctness, coherence, and factuality.
\end{itemize}

\section{Related Work}
\subsection{Structure-Aware Retrieval-Augmented Generation}

Retrieval-Augmented Generation (RAG) enhances LLMs in knowledge-intensive tasks by retrieving external evidence \citep{lewis2020retrieval}. However, conventional RAG methods typically treat retrieved chunks as isolated and flat sequences, overlooking their structural interconnections. To mitigate this, recent research has explored structure-aware variants of RAG. Graph-based methods \citep{nigatu-etal-2025-mrakl, hu-etal-2025-grag, wu-etal-2025-medical, zhu-etal-2025-knowledge,zhou2026arkanswercentricretrievertuning} such as GraphRAG \citep{edge2024local} and KG-RAG \citep{sanmartin2024kg} organize evidence into knowledge graphs, while subsequent work has improved retrieval by simulating human memory mechanisms \citep{gutierrez2024hipporag, gutierrez2025from} or enriching graph semantics \citep{10.1145/3701716.3715240}. Other approaches construct structured subgraphs for coherence \citep{mavromatis-karypis-2025-gnn, li2025simple}, or employ alternative formats like hierarchical graphs \citep{wang2025archrag, huang2025retrieval}, semantic chunking \citep{wang-etal-2025-document, qu-etal-2025-semantic, zhao-etal-2025-moc}, trees \citep{sarthi2024raptor}, and tables \citep{lin2025srag}. More adaptive strategies dynamically select structures based on context \citep{li2025structrag}. Despite these advances, most efforts emphasize surface-level associations (\textit{e.g.,} linking entities) while largely overlooking the rhetorical structure that governs causal flow, evidence presentation, and conclusion formulation. This hinders logical depth and discourse coherence, which our work seeks to address.

\subsection{Rhetorical Structure Theory for Text Generation}
Rhetorical Structure Theory (RST; \cite{mann1987rhetorical, mann1988rhetorical}) is a discourse framework that models hierarchical dependencies and rhetorical relations among Elementary Discourse Units (EDUs). It distinguishes between \textit{nucleus} and \textit{satellite} units, connected by relations such as \textit{Elaboration}, \textit{Causality}, and \textit{Contrast}, forming tree structures that reflect communicative intent. Foundational work \citep{marcu1997discourse, marcu1999decision, mann1987rhetorical, bhatia-etal-2015-better, hayashi-etal-2016-empirical} has established strong correlations between rhetorical structure and human text planning \citep{adewoyin-etal-2022-rstgen}. Later studies have leveraged RST by converting trees into dependency graphs or imposing structural constraints to improve coherence and consistency in neural generation models \citep{chistova-2023-end, zeldes-etal-2025-erst, chistova-2024-bilingual, maekawa-etal-2024-obtain}. More recent efforts have integrated RST into LLMs to improve cross-sentence reasoning and enhance both structural integrity and interpretability of generated outputs \citep{pu-etal-2023-incorporating, pu-demberg-2024-rst}. Compared with shallow discourse markers or sentence-level connectives, the present work extends RST modeling to the RAG setting by explicitly encoding the deeper structure of retrieved passages and highlighting the importance of hierarchical structure.

\section{Proposed Method}

\begin{figure*}[t]
  \centering 
  \includegraphics[width=1\textwidth]{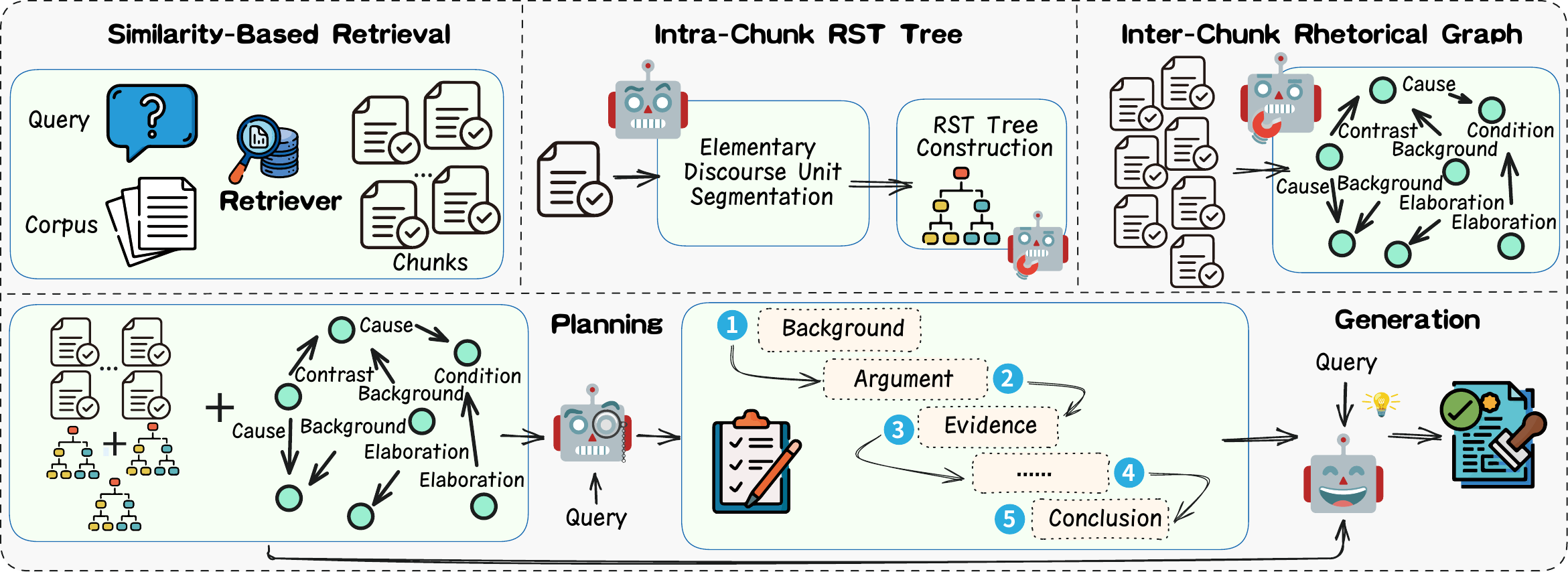}
  \caption{The Disco-RAG pipeline: Starting from passage retrieval (providing context), then intra-chunk RST tree parsing (capturing local discourse), inter-chunk rhetorical graph construction (modeling global discourse), rhetorical planning (blueprint generation), and answer generation (producing the final output).}
  \label{fig:disco_rag}
\end{figure*}

\paragraph{Method Overview.}
We formalize the standard RAG as a conditional generation problem. Given a query $q$ and a set of Top-$k$ retrieved chunks $\mathcal{C}(q; \mathcal{D}) = \{c_1, c_2, \ldots, c_k\}$ from a corpus $\mathcal{D}$, the output is $y = \arg\max_{y'} P(y' \mid q, \mathcal{C}(q; \mathcal{D}))$, where $P(\cdot)$ denotes the conditional distribution of the answer generator. To overcome the limitations of the retrieval-and-concatenation paradigm, we propose \texttt{Disco-RAG} to augment standard RAG with rhetorical parsing and discourse-aware planning.

As illustrated in \autoref{fig:disco_rag}, our pipeline consists of three main stages. (1) we delve into each chunk $c_i$ to uncover its internal logical hierarchy by constructing an intra-chunk RST tree $t_i$, (2) we zoom out to map the relational landscape across all chunks $\mathcal{C}$ via an inter-chunk rhetorical graph $\mathcal{G}$, and (3) we apply a discourse-driven planning module that devises a blueprint $\mathcal{B}$ based on $\mathcal{T} = t_{i=1}^k$ and $\mathcal{G}$ to guide the final generation process.

We hypothesize that under identical retriever and decoding conditions, explicitly injecting discourse knowledge improves the correctness, coherence, and factual consistency of generated text. Here, rhetorical modeling serves as a \textit{knowledge-level prior}, while planning offers \textit{reasoning-level guidance}, jointly inducing stronger structural biases than standard RAG. The following paragraphs provide a detailed account of each component.

\paragraph{Intra-Chunk RST Tree.}
For each retrieved chunk $c_i$, we construct an RST tree $t_i$ using an LLM-based RST parser $\mathcal{A}$ to model local coherence.\footnote{Prompt is detailed in Appendix \autoref{appendix:prompt_rst}.} Given $c_i$, parser $\mathcal{A}$ jointly performs elementary discourse unit (EDU) segmentation and RST parsing, producing a sequence of EDUs $\{e_{i_1}, \ldots, e_{i_m}\}$, nucleus and satellite role assignments, and rhetorical relations among EDUs. Formally, $c_i \xrightarrow{\mathcal{A}} t_i = (V_i, E_i)$, where $V_i = \{e_{i_1}, \ldots, e_{i_m}\}$ is the set of EDU nodes, $\mathcal{R}$ is the set of rhetorical relations (\textit{e.g.}, \textit{Elaboration}, \textit{Contrast}, and \textit{Cause}), and $E_i \subseteq V_i \times V_i \times \mathcal{R}$ is the set of directed connections labeled with relation types. The symbol $\times$ denotes the \textit{cartesian product}. The top-middle panel of \autoref{fig:disco_rag} shows how EDUs are organized into hierarchical trees.\footnote{Intra-chunk RST trees are constructed offline.}

The RST tree parsing is formalized as $P(t_i \mid c_i; \theta_{\mathcal{A}}) = \prod_{j=1}^m P(e_{i_j} \mid c_i; \theta_{\mathcal{A}}) \cdot \prod_{(u,v)} P(r_{u,v} \mid e_{i_u}, e_{i_v}; \theta_{\mathcal{A}})$, where $P(e_{i_j} \mid c_i)$ signifies the probability of EDU boundary prediction and $u, v \in V_i$ are discourse units, $P(r_{u,v} \mid e_{i_u}, e_{i_v})$ corresponds to the probability of the rhetorical relation between two EDUs, and $\theta_{\mathcal{A}}$ indicates the parameters of the parser. 

\paragraph{Inter-Chunk Rhetorical Graph.}

For all retrieved chunks $\mathcal{C}$, we construct a directed graph $\mathcal{G} = (\mathcal{C}, \mathcal{F})$. The edge set $\mathcal{F} \subseteq \mathcal{C} \times \mathcal{C} \times (\mathcal{R} \cup {\texttt{UNRELATED}})$ encodes rhetorical relations or lack thereof. We adopt a listwise inference strategy, where all retrieved chunks $\mathcal{C}$ are provided to parser $\mathcal{A}$ in a single pass, and $\mathcal{A}$ jointly predicts a set of directed rhetorical relations $\{r_{i,j}\}$ or an \texttt{UNRELATED} label for all chunk pairs.\footnote{Appendix \autoref{appendix:prompt_graph} provides prompt and format details used in inter-chunk relation prediction.}

The rhetorical graph construction is modeled as $P(\mathcal{G} \mid \mathcal{C};\theta_{\mathcal{A}})$. This joint distribution can be factorized over ordered chunk pairs as $P(\mathcal{G} \mid \mathcal{C};\theta_{\mathcal{A}}) = \prod_{i=1}^k \prod_{j=1, j \neq i}^k P(r_{i,j} \mid \mathcal{C};\theta_{\mathcal{A}})$. As shown in the top-right panel of \autoref{fig:disco_rag}, the resulting graph $\mathcal{G}$ serves as a global discourse scaffold, allowing the generator to reason over cross-chunk connections.

\paragraph{Discourse-Driven Planning.}
To move beyond the flat concatenation of retrieved evidence, we introduce a planning module that produces a rhetorically informed blueprint to guide the text generation. This is modeled through a mapping from the input query $q$, retrieved chunks $\mathcal{C}$ together with their RST trees $\mathcal{T}$, and the inter-chunk rhetorical graph $\mathcal{G}$ into a discourse-aware plan $(q, \mathcal{C},\mathcal{T}, \mathcal{G}) \xrightarrow{\mathcal{A}} \mathcal{B}$.

As depicted in the center-bottom panel of \autoref{fig:disco_rag}, the plan $\mathcal{B}$ is dynamically conditioned on the discourse structures and the query.\footnote{Appendix \autoref{appendix:prompt_plan} provides prompt used in discourse-aware planning.} The plan outlines reasoning steps that involve selecting salient content, organizing argumentative flow, and prioritizing supporting evidence.

\paragraph{Discourse-Guided RAG.} 
The final stage of generation is conditioned on four inputs: (1) the original text chunks $\mathcal{C}$; (2) the intra-chunk RST trees $\mathcal{T}$; (3) the inter-chunk rhetorical graph $\mathcal{G}$; and (4) the discourse-aware plan $\mathcal{B}$. The objective is $y = \arg\max_{y'} P\big(y' \mid q, \mathcal{C}, \mathcal{T}, \mathcal{G}, \mathcal{B})$, where $y'$ denotes a candidate output and $y$ refers to the final output that maximizes the conditional probability.\footnote{Appendix \autoref{appendix:prompt_generation} contains the generation prompt.}

\section{Experimental Settings}

\paragraph{Evaluation Datasets.} We evaluate our method on three benchmarks, namely Loong \citep{wang-etal-2024-leave}, ASQA \citep{stelmakh-etal-2022-asqa}, and SciNews \citep{pu-etal-2024-scinews}. The Loong dataset focuses on knowledge-intensive reasoning with Spotlight Locating (Spot.), Comparison (Comp.), Clustering (Clus.), and Chain of Reasoning (Chain.). These tasks are evaluated under varying document lengths, where longer inputs increase evidence fragmentation and reasoning difficulty. ASQA involves long-form question answering and requires models to generate responses that are coherent and factually grounded. SciNews targets long-document lay summarization, where the objective is to rewrite scientific articles into accurate and accessible summaries for general audiences \citep{cachola2025evaluating}.

\paragraph{Automatic Metrics.} To ensure consistency and fair comparison, we follow the official evaluation protocols provided by each dataset's repository \citep{wang-etal-2024-leave, stelmakh-etal-2022-asqa, pu-etal-2024-scinews}. For the Loong dataset \citep{wang-etal-2024-leave, li2025structrag}, we report results using Exact Match (EM) and LLM-based scores. For ASQA \citep{stelmakh-etal-2022-asqa, chang-etal-2025-main}, the evaluation includes EM, ROUGE-L (RL) \citep{lin-2004-rouge}, and DR Score \citep{stelmakh-etal-2022-asqa}. On SciNews, we evaluate with RL, BERTScore \citep{Zhang2020BERTScore}, SARI \citep{xu-etal-2016-optimizing}, and SummaC \citep{laban-etal-2022-summac}. These metrics assess the informativeness, fluency, and factual consistency of generated answers. Detailed descriptions of these metrics are provided in \autoref{evaluation_metrics}.

\paragraph{Implementation Details.} Unless specified otherwise, we use \texttt{Llama-3.1-8B}, \texttt{Llama-3.3-70B}, or \texttt{Qwen2.5-72B} across all modules to instantiate and compare performance at different model scales and families \citep{grattafiori2024llama}.\footnote{\texttt{Llama-3.1-8B}, \texttt{Llama-3.3-70B}, and \texttt{Qwen2.5-72B} are the abbreviated names for \texttt{Llama-3.1-8B-Instruct}, \texttt{Llama-3.3-70B-Instruct}, and \texttt{Qwen2.5-72B-Instruct}.} For embedding and retrieval modules, we utilize \texttt{Qwen3-Embedding-8B} \citep{zhang2025qwen3} with a chunk size of 256 tokens without sliding window, and Top-10 retrieval based on cosine semantic similarity. We run each setting once; we use beam search with a beam width of 3, and fix all retrieval settings across all compared methods.

\paragraph{Selected Baselines.} We compare \texttt{Disco-RAG} against three baseline settings: (1) zero-shot LLMs (\texttt{Llama-3.1-8B}, \texttt{Llama-3.3-70B}, and \texttt{Qwen2.5-72B}) with full input context. (2) standard RAG approach \citep{lewis2020retrieval}\footnote{Prompts for full context generation and standard RAG method are provided in Appendix \autoref{appendix:prompt_full_context} and \autoref{appendix:prompt_standard_rag}.}, where relevant chunks are prepended to the query prior to inference.\footnote{All experiments are training-free (unless otherwise specified) and use only task instructions. All hyperparameters follow the same settings as Disco-RAG.} and (3) previously published results from state-of-the-art RAG (if applicable) systems on the same benchmark.

\section{Results and Analysis}

\paragraph{Main Results.} The experimental results are summarized in \autoref{tab:loong_results}, \autoref{tab:asqa_results}, and \autoref{tab:scinews_results}, which correspond to the Loong, ASQA, and SciNews benchmarks, respectively. Across all benchmarks and evaluation metrics, \texttt{Disco-RAG} consistently delivers stable and substantial improvements over the standard RAG baseline.

\begin{table*}[t]
\centering
\small
\resizebox{\textwidth}{!}{
\begin{tabular}{cccccccccccc}
\toprule
\multirow{2}{*}{\textbf{Condition}} & \multirow{2}{*}{\textbf{Model}}
& \multicolumn{2}{c}{\textbf{Spot.}} 
& \multicolumn{2}{c}{\textbf{Comp.}} 
& \multicolumn{2}{c}{\textbf{Clus.}} 
& \multicolumn{2}{c}{\textbf{Chain.}} 
& \multicolumn{2}{c}{\textbf{Overall}} \\
\cmidrule(lr){3-4}
\cmidrule(lr){5-6}
\cmidrule(lr){7-8}
\cmidrule(lr){9-10}
\cmidrule(lr){11-12}
& & \textbf{LLM Score$_{\uparrow}$} & \textbf{EM$_{\uparrow}$} 
& \textbf{LLM Score$_{\uparrow}$} & \textbf{EM$_{\uparrow}$} 
& \textbf{LLM Score$_{\uparrow}$} & \textbf{EM$_{\uparrow}$} 
& \textbf{LLM Score$_{\uparrow}$} & \textbf{EM$_{\uparrow}$} 
& \textbf{LLM Score$_{\uparrow}$} & \textbf{EM$_{\uparrow}$}
 \\
\midrule

\rowcolor{gray!25}
\multicolumn{12}{c}{\textit{Set 1 (10K--50K Tokens)}} \\

\multirow{3}{*}{\scriptsize\textbf{\textit{{Full Context}}}} 
& Llama-3.1-8B & 55.43 & 0.35 & 56.06 & 0.36 & 47.41 & 0.08 & 65.66 & 0.37 & 56.16 & 0.30 \\
& Qwen2.5-72B & 55.11 & 0.34 & 57.21 & 0.33 & 47.09 & 0.10 & 66.51 & 0.36 & 56.59 & 0.31 \\
& Llama-3.3-70B & 58.82 & 0.44 & 61.33 & 0.35 & 48.15 & 0.11 & \textbf{\textcolor{red}{70.31}} & 0.37 & 59.54 & 0.32 \\
\cdashline{1-12}

\multirow{3}{*}{\scriptsize\textbf{\textit{{Standard RAG}}}} 
& Llama-3.1-8B & 62.61 & 0.32 & 60.61 & 0.26 & 53.61 & 0.08 & 58.76 & 0.32 & 60.08 & 0.25 \\
& Qwen2.5-72B & 63.20 & 0.32 & 61.29 & 0.35 & 54.14 & 0.11 & 64.67 & 0.34 & 61.58 & 0.33 \\
& Llama-3.3-70B & 68.44 & 0.45 & 65.32 & 0.39 & 55.30 & 0.12 & 66.48 & 0.36 & 62.78 & 0.34 \\
\cdashline{1-12}

\multirow{3}{*}{\scriptsize\textbf{\textit{{SOTA Results}}}} 
& RQ-RAG$^\star$ \citep{chan2024rqrag} & 72.31 &\textcolor{red}{\textbf{0.54}}  & 48.16 & 0.05 & 47.44 & 0.07 & 58.96 & 0.25 & 53.51 & 0.17 \\
& GraphRAG$^\star$ \citep{edge2024local} & 31.67 & 0.00 & 27.60 & 0.00 & 40.71 & 0.14 & 54.29 & \textbf{\textcolor{red}{0.43}} & 40.82 & 0.18 \\
& StructRAG \citep{li2025structrag} & \textcolor{blue}{\underline{74.53}} & \underline{\textcolor{blue}{0.47}} & \textcolor{blue}{\underline{75.58}} & \textbf{\textcolor{red}{0.47}} & \textcolor{blue}{\underline{65.13}} & \textbf{\textcolor{red}{0.23}} & 67.84 & 0.34 & \textcolor{blue}{\underline{69.43}} & \textcolor{blue}{\underline{0.35}} \\
\cdashline{1-12}

& Disco-RAG (Llama-3.1-8B) & 73.35 & 0.40 & 73.57 & 0.37 & 64.44 & 0.12 & 68.01 & 0.34 & 69.18 & 0.32 \\
& Disco-RAG (Qwen2.5-72B) & 74.46 & 0.42 & 74.39 & 0.41 & 64.66 & 0.15 & 67.73 & 0.35 & 69.39 & 0.33 \\
& Disco-RAG (Llama-3.3-70B) & \textbf{\textcolor{red}{76.60}} & 0.45 & \textbf{\textcolor{red}{75.65}} & \textcolor{blue}{\underline{0.45}} & \textbf{\textcolor{red}{65.36}} & \textcolor{blue}{\underline{0.17}} & \textcolor{blue}{\underline{68.30}} & \underline{\textcolor{blue}{0.38}} & \textbf{\textcolor{red}{71.00}} & \textbf{\textcolor{red}{0.38}} \\
\midrule

\rowcolor{gray!25}
\multicolumn{12}{c}{\textit{Set 2 (50K--100K Tokens)}} \\

\multirow{3}{*}{\scriptsize\textbf{\textit{{Full Context}}}} 
& Llama-3.1-8B & 51.30 & 0.27 & 42.37 & 0.21 & 38.32 & 0.06 & 44.49 & 0.11 & 43.78 & 0.14 \\
& Qwen2.5-72B & 52.37 & 0.30 & 44.47 & 0.25 & 39.24 & 0.07 & 47.69 & 0.11 & 46.61 & 0.13 \\
& Llama-3.3-70B & 55.27 & 0.34 & 47.93 & 0.26 & 40.05 & 0.08 & 50.08 & 0.10 & 48.24 & 0.17 \\
\cdashline{1-12}

\multirow{3}{*}{\scriptsize\textbf{\textit{{Standard RAG}}}} 
& Llama-3.1-8B & 57.02 & 0.25 & 45.42 & 0.19 & 44.21 & 0.05 & 50.42 & 0.15 & 49.12 & 0.16 \\
& Qwen2.5-72B & 60.13 & 0.26 & 50.64 & 0.20 & 45.17 & 0.05 & 53.28 & 0.16 & 50.33 & 0.17 \\
& Llama-3.3-70B & 60.38 & 0.27 & 53.37 & 0.22 & 45.76 & 0.07 & 56.73 & 0.18 & 53.77 & 0.18 \\
\cdashline{1-12}

\multirow{3}{*}{\scriptsize\textbf{\textit{{SOTA Results}}}} 
& RQ-RAG$^\star$ \citep{chan2024rqrag} & 57.35 & 0.35 & 50.83 & 0.16 & 42.85 & 0.03 & 47.60 & 0.10 & 47.09 & 0.10 \\
& GraphRAG$^\star$ \citep{edge2024local} & 24.80 & 0.00 & 14.29 & 0.00 & 37.86 & 0.00 & 46.25 & 0.12 & 33.06 & 0.03 \\
& StructRAG \citep{li2025structrag} & \textcolor{blue}{\underline{68.00}} & \textbf{\textcolor{red}{0.41}} & 63.71 & \textbf{\textcolor{red}{0.36}} & \textcolor{blue}{\underline{61.40}} & \textcolor{blue}{\underline{0.17}} & 54.70 & 0.19 & 60.95 & 0.24 \\
\cdashline{1-12}

& Disco-RAG (Llama-3.1-8B) & 66.03 & 0.36 & 63.56 & 0.24 & 59.53 & 0.14 & 53.06 & 0.16 & 59.03 & 0.23 \\
& Disco-RAG (Qwen2.5-72B) & 67.17 & 0.36 &  \textcolor{blue}{\underline{64.06}} & \textcolor{blue}{\underline{0.30}} & 60.63 & 0.15 & \textcolor{blue}{\underline{57.22}} & \textcolor{blue}{\underline{0.20}} & \textcolor{blue}{\underline{61.32}} & \textcolor{blue}{\underline{0.25}} \\
& Disco-RAG (Llama-3.3-70B) & \textbf{\textcolor{red}{69.92}} & \textcolor{blue}{\underline{0.39}} & \textbf{\textcolor{red}{64.34}} & \textbf{\textcolor{red}{0.36}} & \textbf{\textcolor{red}{61.67}} & \textbf{\textcolor{red}{0.18}} & \textbf{\textcolor{red}{58.23}} & \textbf{\textcolor{red}{0.22}} & \textbf{\textcolor{red}{63.61}} & \textbf{\textcolor{red}{0.28}} \\
\midrule

\rowcolor{gray!25}
\multicolumn{12}{c}{\textit{Set 3 (100K--200K Tokens)}} \\

\multirow{3}{*}{\scriptsize\textbf{\textit{{Full Context}}}} 
& Llama-3.1-8B & 42.25 & 0.22 & 37.43 & 0.12 & 32.27 & 0.00 & 35.62 & 0.00 & 36.51 & 0.08 \\
& Qwen2.5-72B & 45.47 & 0.29 & 40.13 & 0.13 & 35.29 & 0.01 & 48.47 & 0.01 & 42.01 & 0.10 \\
& Llama-3.3-70B & 47.31 & 0.31 & 41.11 & 0.14 & 35.64 & 0.01 & 49.78 & 0.01 & 42.27 & 0.11 \\
\cdashline{1-12}

\multirow{3}{*}{\scriptsize\textbf{\textit{{Standard RAG}}}} 
& Llama-3.1-8B & 49.22 & 0.21 & 40.24 & 0.03 & 36.04 & 0.00 & 49.05 & 0.00 & 43.42 & 0.06 \\
& Qwen2.5-72B & 50.14 & 0.25 & 41.83 & 0.04 & 40.07 & 0.03 & 49.09 & 0.02 & 44.38 & 0.11 \\
& Llama-3.3-70B & 50.33 & 0.33 & 43.70 & 0.06 & 40.13 & 0.04 & 50.10 & 0.05 & 45.77 & 0.13 \\
\cdashline{1-12}

\multirow{3}{*}{\scriptsize\textbf{\textit{{SOTA Results}}}} 
& RQ-RAG$^\star$ \citep{chan2024rqrag} & 50.50 & 0.13 & 44.62 & 0.00 & 36.98 & 0.00 & 36.79 & 0.07 & 40.93 & 0.05 \\
& GraphRAG$^\star$ \citep{edge2024local} & 15.83 & 0.00 & 27.40 & 0.00 & 42.50 & 0.00 & 43.33 & \textcolor{red}{\textbf{0.17}} & 33.28 & 0.04 \\
& StructRAG \citep{li2025structrag} & \textbf{\textcolor{red}{68.62}} & \textbf{\textcolor{red}{0.44}} & \textcolor{blue}{\underline{57.74}} & \textbf{\textcolor{red}{0.35}} & \textcolor{blue}{\underline{58.27}} & \textbf{\textcolor{red}{0.10}} & 49.73 & 0.13 & \textcolor{blue}{\underline{57.92}} & \textcolor{blue}{\underline{0.21}} \\
\cdashline{1-12}

& Disco-RAG (Llama-3.1-8B) & 60.76 & 0.26 & 55.80 & 0.11 & 53.07 & 0.05 & 50.31 & 0.08 & 56.64 & 0.15 \\
& Disco-RAG (Qwen2.5-72B) & 65.58 & 0.33 & 56.89 & 0.19 & 57.23 & 0.06 & \textcolor{blue}{\underline{51.20}} & 0.13 & 57.14 & 0.18 \\
& Disco-RAG (Llama-3.3-70B) & \textcolor{blue}{\underline{66.37}} & \textcolor{blue}{\underline{0.38}} & \textbf{\textcolor{red}{57.84}} & \textcolor{blue}{\underline{0.28}} & \textbf{\textcolor{red}{58.85}} & \textcolor{blue}{\underline{0.07}} & \textbf{\textcolor{red}{52.17}} & \underline{\textcolor{blue}{0.15}} & \textbf{\textcolor{red}{58.86}} & \textbf{\textcolor{red}{0.22}} \\
\midrule

\rowcolor{gray!25}
\multicolumn{12}{c}{\textit{Set 4 (200K--250K Tokens)}} \\

\multirow{3}{*}{\scriptsize\textbf{\textit{{Full Context}}}} 
& Llama-3.1-8B & 31.79 & 0.12 & 25.37 & 0.06 & 27.87 & 0.00 & 26.76 & 0.00 & 27.82 & 0.04 \\
& Qwen2.5-72B & 34.22 & 0.18 & 28.23 & 0.06 & 28.11 & 0.00 & 28.48 & 0.00 & 30.15 & 0.04 \\
& Llama-3.3-70B & 36.76 & 0.21 & 32.22 & 0.07 & 30.69 & 0.00 & 30.17 & 0.00 & 32.21 & 0.05 \\
\cdashline{1-12}

\multirow{3}{*}{\scriptsize\textbf{\textit{{Standard RAG}}}} 
& Llama-3.1-8B & 40.01 & 0.11 & 31.90 & 0.00 & 32.33 & 0.00 & 29.92 & 0.00 & 33.52 & 0.02 \\
& Qwen2.5-72B & 40.14 & 0.16 & 32.31 & 0.01 & 34.00 & 0.00 & 30.02 & 0.01 & 33.64 & 0.04 \\
& Llama-3.3-70B & 40.27 & \underline{\textcolor{blue}{0.25}} & 34.49 & 0.02 & 36.41 & 0.01 & 31.33 & 0.02 & 35.61 & 0.07 \\
\cdashline{1-12}

\multirow{3}{*}{\scriptsize\textbf{\textit{{SOTA Results}}}} 
& RQ-RAG$^\star$ \citep{chan2024rqrag} & 29.17 & 0.08 & 40.36 & 0.00 & 26.92 & 0.00 & 34.69 & 0.00 & 31.91 & 0.01 \\
& GraphRAG$^\star$ \citep{edge2024local} & 17.50 & 0.00 & 26.67 & 0.00 & 20.91 & 0.00 & 33.67 & \textcolor{red}{\textbf{0.33}} & 23.47 & 0.05 \\
& StructRAG \citep{li2025structrag} & 56.87 & 0.19 & \textcolor{blue}{\underline{55.62}} & \textcolor{red}{\textbf{0.25}}  & 56.59 & 0.00 & 35.71 & 0.05 & 51.42 & \textcolor{blue}{\underline{0.10}} \\
\cdashline{1-12}

& Disco-RAG (Llama-3.1-8B) & 56.68 & 0.19 & 53.92 & 0.12 & \textbf{\textcolor{red}{57.53}} & \textcolor{blue}{\underline{0.02}} & 36.00 & 0.03 & 50.87 & 0.08 \\
& Disco-RAG (Qwen2.5-72B) & \textcolor{blue}{\underline{57.27}} & 0.22 & 54.97 & 0.15 & \textcolor{blue}{\underline{57.40}} & \textcolor{blue}{\underline{0.02}} &\textbf{\textcolor{red}{36.17}} & \textcolor{blue}{\underline{0.06}} & \textcolor{blue}{\underline{54.47}} & \textcolor{blue}{\underline{0.10}} \\
& Disco-RAG (Llama-3.3-70B) & \textbf{\textcolor{red}{57.74}} & \textcolor{red}{\textbf{0.27}} & \textbf{\textcolor{red}{55.81}} & \textcolor{blue}{\underline{0.17}}  & 57.36 & \textbf{\textcolor{red}{0.03}} & 	\textcolor{blue}{\underline{36.06}} & \textcolor{blue}{\underline{0.06}} & \textbf{\textcolor{red}{54.62}} & \textbf{\textcolor{red}{0.11}} \\
\bottomrule
\end{tabular}
}
\caption{Loong benchmark results across four document-length settings. Our method (\texttt{Disco-RAG}) is compared against zero-shot LLMs with full context, standard RAG, and prior SOTA. $^\star$ means that the results are directly taken from \citet{li2025structrag}. We use \textbf{\textcolor{red}{bold red}} to indicate the best results and \textcolor{blue}{\underline{blue underlined text}} to indicate the second-best results.
}
\label{tab:loong_results}
\end{table*}

On the Loong benchmark, \texttt{Disco-RAG} demonstrates clear gains across varying document-length settings. With \texttt{Llama-3.3-70B} as the backbone, our method achieves an LLM Score of 71.00 in Set 1, outperforming standard RAG by 8.22 points. The performance gap becomes more significant in Set 4, where \texttt{Disco-RAG} scores 54.62 compared to 35.61 for standard RAG. Averaged across all four sets, our approach also surpasses the best previously reported training-based method \texttt{StructRAG} (62.07 vs. 60.38).

\begin{table}[t]
\centering
\scriptsize
\setlength{\tabcolsep}{3pt}
\begin{tabular}{lccc}
\toprule
\textbf{Model} & \textbf{EM$_{\uparrow}$} & \textbf{RL$_{\uparrow}$} & \textbf{DR Score$_{\uparrow}$}\\
\midrule
\rowcolor{gray!15}
\multicolumn{4}{c}{Baselines with full context} \\
Llama-3.1-8B & 20.1 & 30.6 & 16.3 \\
Qwen2.5-72B & 21.3 & 31.8 & 17.1 \\
Llama-3.3-70B & 22.7 & 32.9 & 16.8 \\
\midrule
\rowcolor{gray!15}
\multicolumn{4}{c}{Baselines with standard RAG} \\
Llama-3.1-8B & 37.3 & 36.9 & 23.4 \\
Qwen2.5-72B & 37.7 & 37.2 & 23.7 \\
Llama-3.3-70B & 38.2 & 37.2 & 24.1 \\
\midrule
\rowcolor{gray!15}
\multicolumn{4}{c}{SOTA Results} \\
FLARE \citep{jiang-etal-2023-active} & 41.3 & 34.3 & 31.1 \\
Tree of Clarifications \citep{kim-etal-2023-tree} & --- & 39.7 & \textbf{\textcolor{red}{36.6}} \\
Open-RAG \citep{islam-etal-2024-open} & 36.3  & 38.1 & --- \\
ConTReGen \citep{roy-etal-2024-contregen} & 41.2  & --- & 30.3 \\
DualRAG \citep{cheng-etal-2025-dualrag} & --- & 31.7 & --- \\
RAS \citep{jiang2025ras} & --- & 39.1 & --- \\
MAIN-RAG-Mistral-7B \citep{chang-etal-2025-main} & 35.7 & 36.2 & --- \\
MAIN-RAG-Llama3-8B \citep{chang-etal-2025-main} & 39.2 & 42.0 & --- \\
\midrule
\rowcolor{gray!15}
\multicolumn{4}{c}{Ours} \\
Disco-RAG (Llama-3.1-8B) & 40.4 & \textcolor{blue}{\underline{42.2}} & 32.6 \\
Disco-RAG (Qwen2.5-72B) & \textcolor{blue}{\underline{41.8}} & 41.3 & \textcolor{blue}{\underline{33.2}} \\
Disco-RAG (Llama-3.3-70B) & \textbf{\textcolor{red}{42.0}} & \textbf{\textcolor{red}{42.3}} & 32.8 \\
\bottomrule
\end{tabular}
\caption{Performance on the ASQA benchmark. \texttt{Disco-RAG} consistently outperforms standard RAG across all metrics. It also surpasses existing SOTA methods on most dimensions.}
\label{tab:asqa_results}
\end{table}

On ASQA, our method again yields consistent advantages. With \texttt{Llama-3.1-8B}, EM, RL, and DR scores increase from 37.3/36.9/23.4 to 40.4/42.2/32.6, and with \texttt{Llama-3.3-70B}, EM rises to 42.0 and DR to 32.8. Notably, our method outperforms more sophisticated prompting systems, such as \texttt{MAIN-RAG} (42.0 RL) and \texttt{Tree of Clarifications} (39.7 RL), achieving an RL score of 42.3. On the SciNews summarization task, our approach exhibits strong generalization ability. Using \texttt{Llama-3.3-70B}, \texttt{Disco-RAG} obtains 21.11 RL score, 65.67 BERTScore, 44.37 SARI, surpassing both standard RAG and the previous best system \citep{pu-etal-2024-scinews, 10.1162/TACL.a.30}.

\begin{table}[t]
\centering
\small
\setlength{\tabcolsep}{3pt}
\resizebox{\linewidth}{!}{%
\begin{tabular}{lcccc}
\toprule
\textbf{Model} & \textbf{RL$_{\uparrow}$} & \textbf{BERTScore$_{\uparrow}$} 
& \textbf{SARI$_{\uparrow}$} & \textbf{SummaC$_{\uparrow}$} \\
\midrule
\rowcolor{gray!15}
\multicolumn{5}{c}{Baselines with full context} \\
Llama-3.1-8B & 15.33 & 59.27 & 35.43 & 48.31 \\
Qwen2.5-72B & 17.00 & 60.41 & 37.62 & 55.03 \\
Llama-3.3-70B & 17.19 & 61.03 & 37.65 & 54.73 \\
\midrule
\rowcolor{gray!15}
\multicolumn{5}{c}{Baselines with standard RAG} \\
Llama-3.1-8B & 17.12 & 60.35 & 38.01 & 55.26 \\
Qwen2.5-72B & 18.09 & 61.28 & 38.32 & 60.12 \\
Llama-3.3-70B & 18.17 & 61.37 & 37.74 & 60.39 \\
\midrule
\rowcolor{gray!15}
\multicolumn{5}{c}{SOTA Results} \\
RSTformer \cite{pu-etal-2024-scinews} & \textcolor{blue}{\underline{20.12}} & 62.80 & \textcolor{blue}{\underline{41.56}} & --- \\
SingleTurnPlan \cite{liang2024integrating} & 19.68 & --- & --- & --- \\
Plan-Input \cite{10.1162/TACL.a.30} & --- & \textcolor{blue}{\underline{65.32}} & --- & \textbf{\textcolor{red}{72.40}} \\
\midrule
\rowcolor{gray!15}
\multicolumn{5}{c}{Ours} \\
Disco-RAG (Llama-3.1-8B) & 19.25 & 63.47 & 40.25 & 63.35 \\
Disco-RAG (Qwen2.5-72B) & 20.10 & 64.83 & 41.48 & 66.30 \\
Disco-RAG (Llama-3.3-70B) & \textbf{\textcolor{red}{21.11}} & \textbf{\textcolor{red}{65.67}} & \textbf{\textcolor{red}{44.37}} & \textcolor{blue}{\underline{69.48}} \\
\bottomrule
\end{tabular}
}
\caption{Performance on the SciNews dataset. \texttt{Disco-RAG} beats both zero-shot and standard RAG, and often surpasses prior SOTA across multiple metrics.}
\label{tab:scinews_results}
\end{table}

\paragraph{Ablation Studies.}

We perform ablation studies on the Loong benchmark, as summarized in \autoref{tab:ablation}, to assess the contribution of each component in \texttt{Disco-RAG}. We find that the removal of any single module leads to performance degradation. The full model achieves an overall LLM Score of 62.07, which drops to 56.22, 57.10, and 59.75 when the RST tree, rhetorical graph, and planner is removed, respectively. Similarly, the Exact Match metric decreases from 0.24 in the full setting to values ranging from 0.20 to 0.22 across the ablated variants. We also include two generic planning baselines built on standard RAG to isolate the added value of discourse structure modeling beyond planning alone.\footnote{Prompts for these two generic planning baselines are provided in Appendix \autoref{appendix:prompt_retrieve_and_plan} and \autoref{appendix:prompt_plan_and_retrieve}.}

\begin{table*}[h]
\centering
\small
\resizebox{\linewidth}{!}{%
\begin{tabular}{lcccccccccc}
\toprule
\multirow{2}{*}{Method} & \multicolumn{2}{c}{Set 1} & \multicolumn{2}{c}{Set 2} & \multicolumn{2}{c}{Set 3} & \multicolumn{2}{c}{Set 4} & \multicolumn{2}{c}{Overall} \\
\cmidrule(lr){2-3} \cmidrule(lr){4-5} \cmidrule(lr){6-7} \cmidrule(lr){8-9} \cmidrule(lr){10-11}
& LLM Score$_{\uparrow}$ & EM$_{\uparrow}$ 
& LLM Score$_{\uparrow}$ & EM$_{\uparrow}$ 
& LLM Score$_{\uparrow}$ & EM$_{\uparrow}$ 
& LLM Score$_{\uparrow}$ & EM$_{\uparrow}$ 
& LLM Score$_{\uparrow}$ & EM$_{\uparrow}$\\
\midrule
\rowcolor{gray!25}
Disco-RAG (full)     
& 71.00 & 0.38
& 63.61 & 0.28
& 58.86 & 0.22
& 54.62 & 0.11
& 62.07 & 0.24 \\
\quad w/o RST tree       
& 65.45 & 0.34 & 58.41 & 0.22 & 54.90 & 0.14 & 47.63 & 0.07 & 56.22 & 0.20 \\
\quad w/o rhetorical graph 
& 67.80 & 0.33 & 58.87 & 0.24 & 54.04 & 0.15 & 48.16 & 0.10 & 57.10 & 0.21 \\
\quad w/o planning       
& 69.11 & 0.35
& 60.14 & 0.25
& 57.20 & 0.20 
& 50.34 & 0.12 
& 59.75 & 0.22 \\
\hdashline
Standard RAG & 62.78 & 0.34 & 53.77 & 0.18 & 45.77 & 0.13 & 35.61 & 0.07 & 49.33 & 0.17 \\
\quad w/ retrieve-and-plan & 64.05 & 0.35 & 54.92 & 0.18 & 46.11 & 0.14 & 37.22 & 0.07 & 50.64 & 0.18 \\
\quad w/ plan-and-retrieve & 64.62 & 0.35 & 55.38 & 0.19 & 47.82 & 0.14 & 38.08 & 0.08 & 51.38 & 0.18 \\
\bottomrule
\end{tabular}
}                 
\caption{Ablation study of the three modules in \texttt{Disco-RAG} with \texttt{Llama-3.3-70B}. \textit{w/o RST tree} removes intra-chunk modeling, \textit{w/o rhetorical graph} removes inter-chunk modeling, and \textit{w/o planning} removes discourse-aware planning. We additionally report two generic planning baselines built on standard RAG. \textit{retrieve-and-plan} generates a free-form plan conditioned on retrieved chunks before generation, and \textit{plan-and-retrieve} first generates a free-form plan from the query and then performs a retrieval step guided by this plan.}
\label{tab:ablation}
\end{table*}

Among the three components, the RST tree and rhetorical graph prove to be the most critical. In the long-document setting (Set 4), eliminating the RST tree leads to a decrease in LLM Score from 54.62 to 47.63. Similarly, removing the rhetorical graph reduces the score to 48.16, whereas excluding the planner causes a smaller drop to 50.34. These findings imply that while all three modules contribute complementarily, structural modeling within and across chunks plays a central role in aggregating information and maintaining discourse coherence.

\paragraph{Impact of Retrieval Granularity and Noise Robustness.}

\begin{figure*}[t]
  \centering \includegraphics[width=1\textwidth]{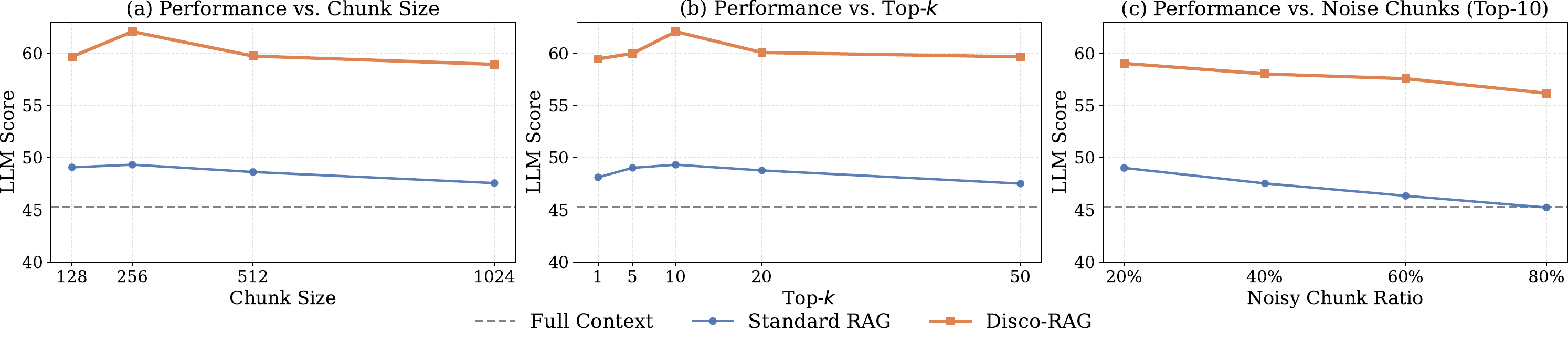}
  \caption{Performance comparison under varying chunk size (a), Top-$k$ value (b), and retrieval noise level (c).}
  \label{fig:overall_chunk_topk_noise}
\end{figure*}

To assess the robustness of \texttt{Disco-RAG} under different retrieval conditions, we execute a series of controlled experiments that manipulate the chunk size of passages, the number of Top-$k$ retrieved chunks, and the proportion of noisy passages. All experiments are conducted on the Loong dataset using \texttt{Llama-3.3-70B} as the backbone model. We maintain identical prompts and decoding configurations across all systems. The evaluation includes two baseline methods, namely the full context setting and the standard RAG framework. Performance is reported using the average LLM Score over four subsets, and the results are visualized in \autoref{fig:overall_chunk_topk_noise}.

Panel (a) of \autoref{fig:overall_chunk_topk_noise} shows that standard RAG performs best at a chunk size of 256 tokens (49.33) but degrades with larger chunks due to the loss of structural coherence. In contrast, \texttt{Disco-RAG} maintains stable performance across all chunk sizes, with scores ranging from 62.07 to 58.91, showing strong robustness to granularity shifts. Panel (b) of \autoref{fig:overall_chunk_topk_noise} shows that while standard RAG peaks at Top-10 and declines with larger $k$ due to accumulating noise, \texttt{Disco-RAG} also performs best at Top-10 but remains robust up to Top-50, showing enhanced capacity to integrate and filter redundant information. Panel (c) of \autoref{fig:overall_chunk_topk_noise} evaluates noise robustness by replacing a fraction of the Top-10 retrieved passages with unrelated content. We randomly replace a proportion of retrieved chunks (\textit{e.g.,} 20\%, 40\%) with irrelevant ones sampled at random from a pool of non-retrieved chunks. The standard RAG exhibits a steep performance drop from 49.33 to 45.23 as noise increases, whereas \texttt{Disco-RAG} retains a score of 56.17, highlighting the structural resilience of our method to retrieval errors.

\paragraph{Impact of Structure Quality and Perturbation Analysis.}
To determine whether the performance gains of \texttt{Disco-RAG} arise from the quality of structural modeling rather than the mere presence of structural cues, we conduct a set of controlled perturbation experiments targeting three core components of our framework. These include intra-chunk RST trees, inter-chunk rhetorical graphs, and discourse-aware plans. For each module, we introduce partial degradations by randomly selecting relation labels, edge directions, or planning steps, and either replacing or removing them. This design ensures that the perturbed structures still retain partial coherence, allowing us to assess how sensitive the model is to incomplete or noisy signals. All experiments are conducted with \texttt{Llama-3.3-70B} under consistent retrieval and decoding conditions to maintain causal interpretability.

\begin{figure*}[t]
  \centering \includegraphics[width=1\textwidth]{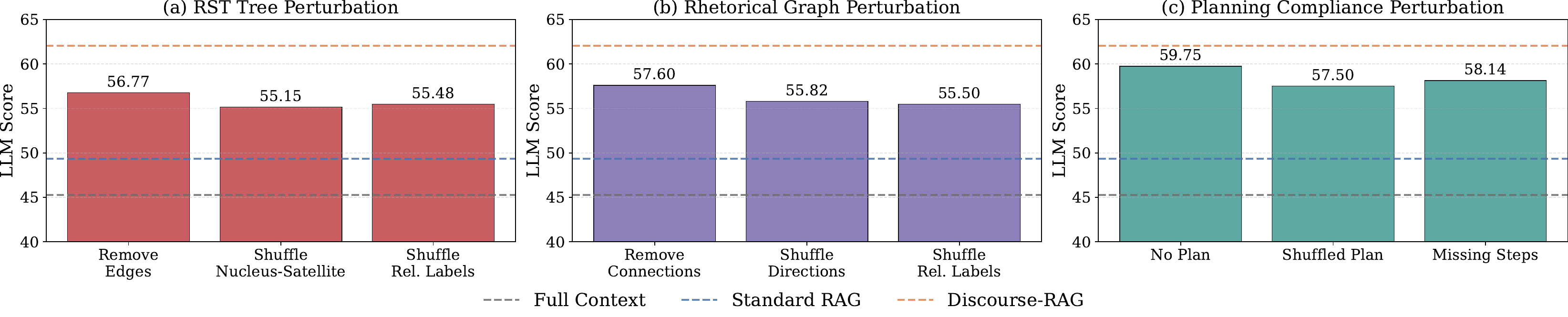}
  \caption{Effect of structural perturbations on performance. Panels (a), (b), and (c) correspond to intra-chunk RST trees, inter-chunk rhetorical graphs, and discourse-aware plans, respectively. Each perturbation involves randomly altering or removing the relevant elements.}
  \label{fig:struct_ablation_results}
\end{figure*}

\autoref{fig:struct_ablation_results} presents the results of the perturbation study. Panel (a) of \autoref{fig:struct_ablation_results} exhibits that perturbing intra-chunk structures leads to a consistent performance decrease. Randomly shuffling a portion of rhetorical relation labels reduces the LLM Score from 62.07 to 55.48. Randomly altering some nucleus–satellite roles lowers the score to 55.15. Removing a randomly selected subtree connection decreases the score to 56.77. Panel (b) of \autoref{fig:struct_ablation_results} presents the effect of modifying rhetorical graphs. Randomly removing some graph connections between chunks reduces the score to 57.60. Randomly flipping the directions of a subset of edges yields 55.82, while replacing some discourse relation labels within the graph gives 55.50. Panel (c) of \autoref{fig:struct_ablation_results} analyzes the degradation of rhetorical plans. Omitting the plan altogether reduces performance to 59.75. Shuffling some of the step sequences causes a decline to 57.50, while removing a subset of steps results in 58.14.

Across all three dimensions, structural perturbations lead to a performance reduction, yet do not eliminate the benefits conferred by structure-aware modeling. Even when exposed to corrupted or incomplete signals, \texttt{Disco-RAG} consistently outperforms both the standard RAG and the full context setting. These results confirm that the observed improvements are not merely due to the inclusion of additional tokens, but instead arise from the model’s capacity to leverage structural signals.

\paragraph{Mixed-Model Deployment.}
Since the structural modules (RST parsing, rhetorical graph construction, and planning) and the final generator are decoupled by design, they can be instantiated with different models. To examine whether a smaller model can serve as a cost-effective structural backbone, we conduct an experiment on the Loong benchmark in which \texttt{Llama-3.1-8B} handles all structural modules while \texttt{Llama-3.3-70B} is used only for the final generation stage. \autoref{tab:mixed_model} reports the results under four configurations.

\begin{table}[h]
\centering
\small
\setlength{\tabcolsep}{4pt}
\resizebox{\linewidth}{!}{%
\begin{tabular}{lccc}
\toprule
\textbf{Configuration} & \textbf{Structurer / Generator} & \textbf{LLM Score$_{\uparrow}$} & \textbf{EM$_{\uparrow}$} \\
\midrule
Standard RAG & 70B / 70B & 49.33 & 0.17 \\
\hdashline
Disco-RAG (all 8B) & 8B / 8B & 58.94 & 0.20 \\
Disco-RAG (8B+70B) & 8B / 70B & 60.52 & 0.22 \\
Disco-RAG (all 70B) & 70B / 70B & 62.07 & 0.24 \\
\bottomrule
\end{tabular}}
\caption{Mixed-model results on the Loong benchmark. \textit{Structurer} refers to the model used for RST parsing, rhetorical graph construction, and planning; \textit{Generator} refers to the model used for final answer generation. 8B = \texttt{Llama-3.1-8B}; 70B = \texttt{Llama-3.3-70B}.}
\label{tab:mixed_model}
\vspace{-5pt}
\end{table}

The mixed configuration (8B structurer + 70B generator) achieves an LLM Score of 60.52, recovering the majority of the gain obtained by the full 70B setting (62.07) and substantially outperforming standard RAG (49.33). Even the all-8B variant reaches 58.94, indicating that smaller models can produce discourse structures that still benefit downstream generation. The remaining gap between the 8B+70B and all-70B settings (1.55 LLM Score) suggests that higher-quality structural signals from larger models still contribute incrementally, yet the cost reduction from offloading structural inference to an 8B model may be preferable in resource-constrained deployments. These results confirm that the modular architecture of \texttt{Disco-RAG} supports flexible allocation of model capacity across components without forfeiting the core benefits of discourse-aware generation.

\paragraph{Effect of Supervised Fine-Tuning.}
We examine how supervised fine-tuning interacts with discourse-aware modeling on the SciNews summarization benchmark. Starting from \texttt{Llama-3.3-70B}, we fine-tune the generator on the SciNews training split with a standard sequence-to-sequence summarization objective and test using the RAG setting under three conditions. In the end-to-end baseline, the model is trained using only the raw document-summary pairs without any discourse inputs. In the second setting, the model is trained in the same way, but at test time, we augment the inputs with the intra-chunk RST trees, inter-chunk rhetorical graphs, and discourse-aware plans produced by \texttt{Disco-RAG}. In the third setting, both training and inference use the discourse-enriched inputs so that the model can adapt its parameters to the structural signals. For comparison, we also include the original training-free \texttt{Disco-RAG} system that conditions generation on discourse structures via prompting without parameter updates.

\begin{table}[h]
\centering
\small
\setlength{\tabcolsep}{6pt}
\resizebox{\linewidth}{!}{%
\begin{tabular}{lcc}
\toprule
\textbf{Method} & \textbf{RL$_{\uparrow}$} & \textbf{SummaC$_{\uparrow}$} \\
\midrule
End-to-end SFT (no discourse) & 20.3 & 66.8 \\
\hdashline
Disco-RAG (training-free) & 21.1 & 69.5 \\
SFT with test time discourse & 22.8 & 72.3 \\
SFT with train and test discourse & 23.3 & 74.0 \\
\bottomrule
\end{tabular}}
\caption{Impact of supervised fine-tuning (SFT) and discourse conditioning.}
\label{tab:training_scinews}
\end{table}

All systems share the same retrieval pipeline and decoding configuration, and we report RL and SummaC on the SciNews test set. \autoref{tab:training_scinews} shows that naive end-to-end fine-tuning improves over the zero-shot standard RAG baselines but remains behind the training-free \texttt{Disco-RAG}. When discourse structures are provided at test time, the fine-tuned model surpasses \texttt{Disco-RAG}, indicating that structural guidance and parameter adaptation bring complementary benefits. When discourse structures are incorporated during both training and inference, we observe further gains in both RL and SummaC. These results confirm that our discourse-aware framework is orthogonal to model training and that injecting discourse information can consistently enhance performance on top of supervised fine-tuning.

\paragraph{Human Evaluation.}
We conduct a human evaluation on the SciNews dataset. We randomly sample 15 test articles and ask three graduate students with computer science backgrounds to rate four anonymized systems, namely the full-context LLM without retrieval, the standard RAG baseline, our \texttt{Disco-RAG} model, and human-written references. Following the protocol of \citet{pu-etal-2024-scinews}, human raters read each article together with four shuffled summaries and assign scores on a three-point Likert scale along four dimensions, \textit{Relevance}, \textit{Simplicity}, \textit{Conciseness}, and \textit{Faithfulness}, where higher values indicate better quality. We measure inter-rater agreement using Fleiss' $\kappa$ and obtain average values of 0.73, 0.65, 0.66, and 0.68 on the four dimensions, indicating substantial consistency among annotators. \autoref{tab:human_eval} reports the average scores across all annotated samples. Detailed instructions for the human raters are provided in \autoref{human_evaluation_guideline}.

\begin{table}[h]
\centering
\small
\setlength{\tabcolsep}{2pt}
\resizebox{\linewidth}{!}{%
\begin{tabular}{lcccc}
\toprule
\textbf{System} & \textbf{Relevance$_{\uparrow}$} & \textbf{Simplicity$_{\uparrow}$} & \textbf{Conciseness$_{\uparrow}$} & \textbf{Faithfulness$_{\uparrow}$} \\
\midrule
Full Context & 1.65 & 1.98 & 1.52 & 1.45 \\
Standard RAG & 1.87 & 2.12 & 1.60 & 1.67 \\
\texttt{Disco-RAG} & 2.40 & 2.43 & 2.27 & 2.53 \\
\hdashline
Human Reference & 2.89 & 2.63 & 2.48 & 2.88 \\
\bottomrule
\end{tabular}}
\caption{Average human ratings on SciNews. Scores are computed on a three-point Likert scale, and higher values indicate better performance.}
\label{tab:human_eval}
\end{table}

\autoref{tab:human_eval} suggests that \texttt{Disco-RAG} improves perceived answer quality over both full context and standard RAG systems, with considerable gains in \textit{Faithfulness} and \textit{Conciseness}. Human-written references remain the strongest overall according to annotators, which indicates that there is still room for future model development, but the ranking of neural systems in human evaluation is consistent with the trends observed in automatic metrics and supports the benefits of discourse-aware retrieval-augmented generation. Further discussion of the parsing evaluation, shallow discourse marker analysis, significance testing, qualitative case studies, and LLM usage can be found in \autoref{appendix:parser}, \autoref{appendix:shallow}, \autoref{appendix:significance}, \autoref{appendix:case_study}, and \autoref{use_of_llms}, respectively.

\section{Conclusion}
In this study, we tackle the absence of discourse structure modeling in existing RAG approaches by presenting \texttt{Disco-RAG}. Grounded in Rhetorical Structure Theory, our approach constructs both local hierarchies and global discourse representations over retrieved evidence and leverages them to derive a high-level blueprint that guides the reasoning process of the language model. Experimental results demonstrate that \texttt{Disco-RAG} achieves considerable gains across multiple knowledge-intensive QA and summarization tasks, surpassing previous state-of-the-art methods without in-domain fine-tuning. Ablation studies validate the complementary contributions of each structural component. Taken together, these findings highlight structured discourse modeling as a promising direction for advancing retrieval-augmented generation systems.

\section*{Ethical Considerations}
All datasets used in this work are publicly available, and we follow the original licenses and usage policies. Our pipeline operates entirely on de-identified text without collecting or inferring personal identities or sensitive attributes, and all intermediate artifacts, such as discourse structures and plans, are derived solely from these corpora rather than live user queries or proprietary logs. Human evaluators participate voluntarily and are appropriately compensated, while care is taken to avoid exposing annotators to harmful content beyond what already exists in the datasets. Large language models are used as backbone retrievers/generators and as assistive tools for discourse parsing and language refinement (as noted in \autoref{use_of_llms}), but they do not replace the authors in methodological design or result interpretation. We also comply with \href{https://www.aclweb.org/adminwiki/index.php/ACL_Policy_on_Publication_Ethics}{ACL Policy on Publication Ethics}, and we caution against applying our system in high-stakes environments without additional safeguards and human oversight.

\section*{Limitations}
\paragraph{Data.} Our experiments apply three publicly available benchmarks, Loong, ASQA, and SciNews. These datasets provide a good basis for evaluating long-context reasoning, but we do not conduct a dedicated analysis of potential biases in their content or label distributions. As a result, the behavior of \texttt{Disco-RAG} across genres, languages, or data collection processes that differ substantially from these benchmarks remains an open question, and extending our analysis to different corpora is a direction for future work.

\paragraph{Model.} We instantiate our framework with three open-source large language models, including \texttt{Llama-3.1-8B}, \texttt{Llama-3.3-70B}, and \texttt{Qwen2.5-72B}, together with a fixed retriever. The consistent gains observed across these backbones suggest that the proposed discourse mechanism is not tied to a specific model, yet we do not systematically explore alternative architectures, parameter scales, or decoding strategies. In practice, our framework assumes that the backbone models provide basic discourse understanding and long-context processing, and we expect that future improvements in foundation models can be incorporated with minimal changes to the overall design so that \texttt{Disco-RAG} remains largely decoupled from specific language models. Further work is also needed to understand how \texttt{Disco-RAG} behaves with smaller or more specialized models and under tighter computational constraints.

\paragraph{Automated Evaluation.} We evaluate models using a combination of Exact Match, ROUGE-L, DR Score, BERTScore, SARI, SummaC, and an LLM-based metric for Loong that relies on \texttt{GPT-4-turbo-2024-04-09}. This suite covers multiple aspects of quality and has been adopted in prior work \cite{pu-simaan-2022-passing, pu-etal-2022-two, pu-demberg-2023-chatgpt, li2026se,zhang2025rediscoveringentropyregularizationadaptive,hu2025stepdeepresearchtechnicalreport,ruan2026aorchestra}, while each metric has known limitations, and the LLM-based judge may inherit biases or topic preferences from its own training data. Reported numbers should therefore be interpreted as indicative rather than exhaustive, and future work could benefit from more fine-grained evaluation protocols and larger-scale human studies.

\paragraph{Efficiency.} Compared with standard RAG, \texttt{Disco-RAG} introduces additional token consumption and latency because it requires additional LLM calls for discourse parsing and rhetorical planning. Deploying our method in latency-sensitive or large-scale applications will therefore require engineering optimizations such as caching and reusing discourse structures, batching structural queries, or distilling lighter parsers and planners, and there remains an inherent trade-off between structural richness and runtime efficiency.

\section*{Acknowledgements}
Dongqi Liu acknowledges support from the European Research Council (ERC) under the European Union's Horizon 2020 Research and Innovation Programme (Grant Agreement No. 948878). Dongqi Liu also thanks Vera Demberg and Mirella Lapata for valuable discussions that contributed to shaping the ideas of this work. All authors of the present work are grateful to the anonymous reviewers and area chairs for their exceptionally detailed and helpful feedback.

\begin{figure}[h] 
\centering
\includegraphics[width=0.75\columnwidth]{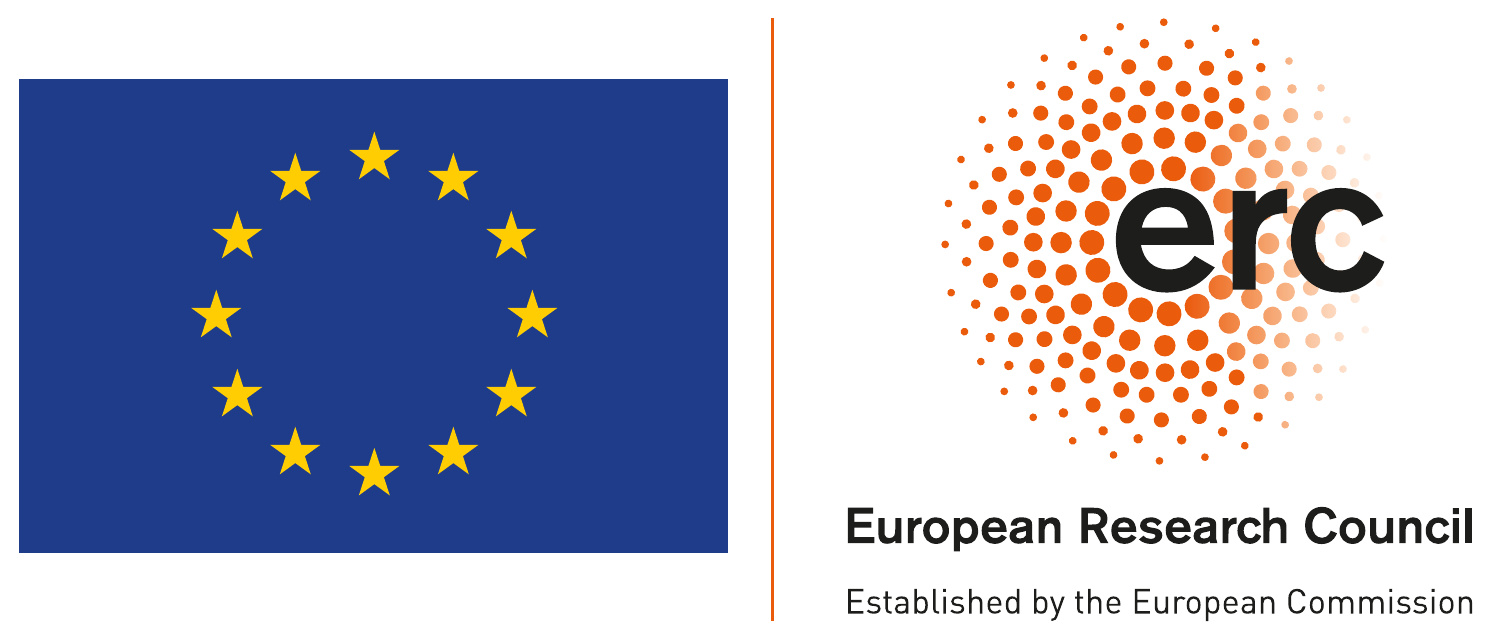}
\end{figure}

\bibliography{custom}

\appendix

\section{Details of Evaluation Metrics}
\label{evaluation_metrics}

\paragraph{For the Loong dataset.} We report two evaluation metrics. The first is Exact Match (EM), which is a strict measure of the percentage of model predictions that exactly match the ground truth answers. It is a binary measure that assigns a score of one for a perfect match and zero otherwise. The second metric is the LLM Score \citep{wang-etal-2024-leave}, ranging from 0 to 100. Following the protocol introduced by the dataset authors, we employ \texttt{GPT-4-turbo-2024-04-09} as an automated evaluator to rate the overall quality of generated responses. Unlike EM, which captures only factual correctness, the LLM Score provides a holistic evaluation by jointly considering comprehensiveness, clarity, and adherence to instructions, thereby offering a more integrated assessment across multiple dimensions of quality.

\paragraph{For the ASQA dataset.} We adopt the standard evaluation suite. The first is Exact Match (EM), defined as before. The second is ROUGE-L \citep{lin-2004-rouge}, an evaluation metric based on the Longest Common Subsequence (LCS). It measures the n-gram overlap between prediction and reference by identifying the longest sequence of words that occurs in both while preserving word order, thereby evaluating the coverage of key information. Given a predicted text $\hat{y}_i$ and a reference text $y_i$, let $LCS(\hat{y}_i, y_i)$ denote the length of their longest common subsequence. The ROUGE-L recall, precision, and F1 are defined as:

\begin{equation}
R_L = \frac{LCS(\hat{y}_i, y_i)}{|y_i|}
\end{equation}
\begin{equation}
P_L = \frac{LCS(\hat{y}_i, y_i)}{|\hat{y}_i|}
\end{equation}
\begin{equation}
F_L = \frac{(1+\beta^2) \cdot R_L \cdot P_L}{R_L + \beta^2 \cdot P_L}
\end{equation}

where $|y_i|$ and $|\hat{y}_i|$ are the lengths of the reference and predicted texts, respectively, and $\beta$ is set to one by default to balance recall and precision. In our experiments, we report ROUGE-L F1 (RL).

The third metric is the Disambiguation Recall (DR) Score \citep{stelmakh-etal-2022-asqa}, which is specifically designed for ASQA to evaluate whether a prediction covers all possible disambiguated answers present in the reference set. While ROUGE-L cannot distinguish between two fluent but semantically divergent answers, the DR score explicitly evaluates coverage across multiple reference answers. A higher DR score indicates that the generated response captures a larger fraction of the possible interpretations of an ambiguous question. Given multiple reference answers $\mathcal{Y}_i = \{y_i^{(1)}, y_i^{(2)}, \dots, y_i^{(k_i)}\}$ for a query and a generated answer $\hat{y}_i$, the instance-level DR score is defined as:

\begin{equation}
\text{DR}_i = \frac{1}{|\mathcal{Y}_i|} \sum_{j=1}^{|\mathcal{Y}_i|}
\mathbf{1}\big[\, y_i^{(j)} \subseteq \hat{y}_i \,\big]
\end{equation}

where $\mathbf{1}[\cdot]$ is an indicator function equal to one if the predicted answer includes the content of a reference answer $y_i^{(j)}$, and zero otherwise. The overall DR score across $N$ queries is defined as:

\begin{equation}
\text{DR} = \frac{1}{N} \sum_{i=1}^{N} \text{DR}_i.
\end{equation}

\paragraph{For the SciNews dataset.} We focus on summarization quality using four metrics. The first is ROUGE-L, as defined above. The second is BERTScore \citep{Zhang2020BERTScore}, which computes semantic similarity between prediction and reference using contextual embeddings from a pre-trained BERT model. The third is SARI \citep{xu-etal-2016-optimizing}, which assesses the quality of simplification by comparing system outputs against both the source text and the reference texts. SARI explicitly measures the precision and recall of words that are added, deleted, and kept. For a source sentence $s_i$, a prediction $\hat{y}_i$, and a set of reference simplifications $\mathcal{Y}_i = \{y_i^{(1)}, \dots, y_i^{(k_i)}\}$, SARI is defined as:

\begin{equation}
\text{SARI} = \frac{1}{3} \Big( \text{Add}_{F_1} + \text{Keep}_{F_1} + \text{Del}_{F_1} \Big)
\end{equation}

where $\text{Add}_{F_1}$, $\text{Keep}_{F_1}$, and $\text{Del}_{F_1}$ denote the F1 scores for added, kept, and deleted n-grams relative to both the source and the reference sets. The fourth metric is SummaC \citep{laban-etal-2022-summac}, a model-based measure of factual consistency. SummaC can be used to determine whether a generated summary is entailed by its source document and detects unsupported or hallucinated content, which is essential for ensuring the reliability of generated text.

\section{Details of Baselines}
Here we describe the baselines used for comparison: 
\begin{itemize}[leftmargin=8pt,itemsep=1pt,topsep=1pt,parsep=1pt]
    \item \textbf{Standard RAG} \citep{lewis2020retrieval} We implement the standard retrieval-augmented generation framework, where a retriever (\texttt{Qwen3-Embedding-8B}) retrieves relevant documents and a generator (\texttt{Llama-3.1-8B}, \texttt{Llama-3.3-70B} or \texttt{Qwen2.5-72B}) produces the final answer conditioned on the retrieved context.
    \item \textbf{GraphRAG} \citep{edge2024local} augments retrieval with a graph-based knowledge representation by constructing a semantic knowledge graph from retrieved passages. It leverages community detection to capture global structures and integrates graph contexts into generation, enabling more accurate and coherent reasoning across documents.
    \item \textbf{RQ-RAG} \citep{chan2024rqrag} refines queries through explicit rewriting, decomposition, and disambiguation before retrieval. It trains LLMs end-to-end on a curated dataset with search-augmented supervision, enabling dynamic query refinement and improving both single-hop and multi-hop QA by learning to search only when needed.
    \item \textbf{FLARE} \citep{jiang-etal-2023-active} actively decides when and what to retrieve during generation by predicting upcoming sentences and using them as queries to fetch additional documents whenever low-confidence tokens appear.
    \item \textbf{Tree of Clarifications} \citep{kim-etal-2023-tree} addresses ambiguous questions by recursively constructing a tree of disambiguated questions with retrieval-augmented few-shot prompting, pruning unhelpful branches through self-verification, and generating a long-form answer that covers all valid interpretations.
    \item \textbf{Open-RAG} \citep{islam-etal-2024-open} enhances retrieval-augmented reasoning with open-source LLMs by transforming a dense model into a parameter-efficient sparse mixture-of-experts, combining contrastive learning against distractors with hybrid adaptive retrieval.
    \item \textbf{ConTReGen} \citep{roy-etal-2024-contregen} employs a context-driven, tree-structured retrieval framework for open-domain long-form text generation. It performs top-down planning to recursively decompose a query into sub-questions for in-depth retrieval, followed by bottom-up synthesis to integrate information from leaf nodes to the root.
    \item \textbf{DualRAG} \citep{cheng-etal-2025-dualrag} introduces a dual-process framework for multi-hop QA, consisting of Reasoning-augmented Querying (RaQ), which identifies knowledge gaps and formulates targeted queries, and progressive Knowledge Aggregation (pKA), which filters and structures retrieved information into a coherent knowledge outline.
    \item \textbf{RAS} \citep{jiang2025ras} interleaves iterative retrieval planning with dynamic construction of query-specific knowledge graphs. It converts retrieved text into factual triples, incrementally builds a structured graph, and conditions generation on the evolving graph. 
    \item \textbf{MAIN-RAG} \citep{chang-etal-2025-main} is a training-free framework that employs three LLM agents to collaboratively filter and rank retrieved documents. It introduces an adaptive judge bar that dynamically adjusts relevance thresholds based on score distributions, effectively reducing noisy retrievals while preserving relevant information.
    \item \textbf{StructRAG} \citep{li2025structrag} introduces hybrid information structurization for knowledge-intensive reasoning. It employs a hybrid structure router to select the optimal structure type (e.g., table, graph, catalogue), a scattered knowledge structurizer to transform raw documents into structured knowledge, and a structured knowledge utilizer to decompose complex questions and infer accurate answers based on the structured representation. 
\end{itemize}

\section{Parsing Evaluation}
\label{appendix:parser}
To assess the parser quality on our framework, we evaluate the LLM-based parser used in \texttt{Disco-RAG} on the RST-DT benchmark following the evaluation protocol of \citet{maekawa-etal-2024-obtain}. We compare a fine-tuned RST parser from \citet{maekawa-etal-2024-obtain} with our zero-shot parser instantiated with \texttt{Llama-3.3-70B}. Both models are evaluated on span F1, nuclearity F1, and relation F1 using the official data splits and scoring scripts of the benchmark, and the results are summarized in \autoref{tab:parser_choice}. Our zero-shot parser attains competitive scores that are close in nuclearity and relation prediction and somewhat lower in span prediction, which still reflects reasonable sensitivity to rhetorical semantics without any task-specific tuning.

\begin{table}[h]
\centering
\setlength{\tabcolsep}{6pt}
\resizebox{\linewidth}{!}{%
\begin{tabular}{lcccc}
\toprule
\textbf{Model} & \textbf{Setting} & \textbf{Span F1} & \textbf{Nuclearity F1} & \textbf{Relation F1} \\
\midrule
\citet{maekawa-etal-2024-obtain} & Supervised & 79.8 & 70.4 & 60.0 \\
Our Parser & Unsupervised & 70.4 & 63.1 & 58.6 \\
\bottomrule
\end{tabular}}
\caption{Evaluation of the RST parser on the RST-DT benchmark following the protocol of \citet{maekawa-etal-2024-obtain}.}
\label{tab:parser_choice}
\end{table}

Note that parser development is not the primary focus of \texttt{Disco-RAG}, and these results indicate that the zero-shot LLM parser provides a reasonable structural signal for downstream reasoning. Moreover, the fine-tuned parser only generates output with a specific format and cannot complete the rhetorical graph prediction between chunks, while the zero-shot parser provides such flexibility.

We further conduct a case study to examine whether the parser outputs are acceptable for the downstream task when gold annotations are unavailable for our benchmarks. For each of Loong, ASQA, and SciNews, we randomly select 10 instances from the test set and run our pipeline to obtain intra-chunk RST trees, inter-chunk rhetorical graphs, and discourse-aware plans for the retrieved evidence. We then ask three human annotators to judge two questions with binary labels. The first question evaluates \textit{whether the predicted discourse structures are broadly acceptable}, meaning that they capture major relations within and between chunks even if they are not perfectly accurate in every detail. The second question evaluates \textit{whether the discourse-aware plan is acceptable}, meaning that it organizes the answer in a reasonable order and reflects the main evidence required by the query.

Across the sampled instances, the average acceptability rates are 0.72 for intra-chunk discourse trees, 0.80 for inter-chunk rhetorical graphs, and 0.93 for discourse-aware plans. The inter-annotator agreement measured by Fleiss' $\kappa$ is 0.71 for intra-chunk discourse trees, 0.73 for inter-chunk rhetorical graphs, and 0.86 for discourse-aware plans, indicating high consistency among annotators. These results suggest that the parsing outputs and the plans provide usable discourse signals for our framework, and we expect that improved parsing performance would further enhance the reliability of discourse structures and thereby support additional gains in answer quality.

\section{Comparison with Shallow Discourse Markers}
\label{appendix:shallow}
We conduct a study on the Loong to assess whether shallow discourse cues alone can provide comparable benefits to full RST-based modeling. To this end, we design a marker-based variant that constructs inter-chunk links using explicit discourse markers without applying EDU segmentation. We adopt a listwise inference strategy and provide all retrieved chunks to \texttt{Llama-3.3-70B} in a single pass, which jointly predicts discourse a marker for each ordered chunk pair based on connective cues such as \textit{however}, \textit{but}, \textit{although}, \textit{in contrast}, \textit{therefore}, \textit{because}, \textit{as a result}, and \textit{meanwhile}.\footnote{The prompt used for shallow discourse marker inference is provided in Appendix \autoref{appendix:prompt_shallow_marker}.} 

\begin{table}[h]
\centering
\small
\setlength{\tabcolsep}{6pt}
\resizebox{\linewidth}{!}{%
\begin{tabular}{lcc}
\toprule
\textbf{Method} & \textbf{LLM Score$_{\uparrow}$} & \textbf{Exact Match$_{\uparrow}$} \\
\midrule
Standard RAG & 49.33 & 0.17 \\
\quad w/ Discourse Markers & 50.41 & 0.20 \\
\hdashline
Disco-RAG & 62.07 & 0.24 \\
\bottomrule
\end{tabular}}
\caption{Comparison of standard RAG, a shallow discourse marker variant, and \texttt{Disco-RAG} on the Loong benchmark with \texttt{Llama-3.3-70B}.}
\label{tab:shallow_markers}
\end{table}

\autoref{tab:shallow_markers} compares three configurations, namely standard RAG, a shallow variant that augments standard RAG with discourse markers, and the full \texttt{Disco-RAG} model. The marker-based system improves LLM Score from 49.33 to 50.41 and Exact Match from 0.17 to 0.20. However, these gains remain modest compared with the full discourse-aware setting, where \texttt{Disco-RAG} reaches 62.07 LLM Score and 0.24 Exact Match under the same conditions.

\section{Significance Testing}
\label{appendix:significance}
To assess whether the improvements of \texttt{Disco-RAG} over standard RAG are statistically reliable under the same backbone model and decoding configuration, we conduct paired t-tests on metric scores for every benchmark, every backbone, and every automatic metric. For human evaluation, we apply the same paired t-test on the instance-level average ratings across the three annotators for each criterion. Across all evaluation settings reported in the paper, \texttt{Disco-RAG} is significantly better than standard RAG with \(p < 0.05\).

\section{Case Studies}
\label{appendix:case_study}

We present qualitative case studies to illustrate the effectiveness of \texttt{Disco-RAG} compared to standard RAG. The three cases are shown in \autoref{fig:case_study_1}, \autoref{fig:case_study_2}, and \autoref{fig:case_study_3}.

\section{Use of Large Language Models}
\label{use_of_llms}
In preparing this paper, we use \texttt{GPT-5.2} as a writing assistant for language polishing, grammar correction, and stylistic refinement. The model is not involved in the research ideation, methodology design, experiments, or result interpretation. The authors take full responsibility for the content of the manuscript, including any parts generated with the assistance of \texttt{GPT-5.2}. In accordance with conference policy, we confirm that the LLM is not an author of this work and does not bear responsibility for its scientific claims.

\begin{figure*}[h]
\centering
\begin{tcolorbox}[
  colback=gray!5!white,
  colframe=gray!40!black,
  width=\linewidth,
  boxrule=0.5pt,
  arc=2pt,
  outer arc=2pt,
  title=Case Study 1,
]
\paragraph{Query.} \textit{When did The Lion King debut on Broadway?}

\paragraph{Retrieved Chunks.}

\begin{verbatim}
[Chunk 1] The Lion King is a stage musical ... It debuted on July 8, 1997, 
in Minneapolis, Minnesota at the Orpheum Theatre ...

[Chunk 2] ... The musical premiered on Broadway at the New Amsterdam Theatre 
in previews on October 15, 1997 ...

[Chunk 3] ... The official Broadway opening was on November 13, 1997, 
at the New Amsterdam Theatre ...

[Chunk 4] ... On June 13, 2006, the Broadway production moved to the Minskoff Theatre, 
where it is still running ...
\end{verbatim}

\paragraph{Standard RAG Output.}
\begin{quote}
The Lion King debuted on Broadway in 1997. It was first performed in Minneapolis before moving to Broadway, where it became one of the most popular shows. 
\end{quote}

\paragraph{Disco-RAG Output.}
\begin{quote}
The Lion King opened on Broadway more than once. It premiered in previews at the New Amsterdam Theatre on October 15, 1997, and officially opened on November 13, 1997. Later, on June 13, 2006, the production moved to the Minskoff Theatre, where it continues to run. 
\end{quote}

\paragraph{Reference Answer.}
\begin{quote}
The Lion King opened on Broadway more than once. It premiered on Broadway at the New Amsterdam Theatre in previews on October 15, 1997, then officially opened on Broadway on November 13, 1997. On June 13, 2006, the Broadway production moved to the Minskoff Theatre, another Broadway theater, where it is still running after more than 9,000 performances.
\end{quote}

\paragraph{Remark.} 
Standard RAG produces a vague answer that misses key details about the Broadway timeline, providing only the year 1997. In contrast, \texttt{Disco-RAG} integrates evidence across all relevant chunks, capturing both the preview and official opening dates as well as the later theater relocation.
\end{tcolorbox}
\caption{Case study comparing standard RAG and \texttt{Disco-RAG} on the query ``When did The Lion King debut on Broadway?''. Our method captures both the preview and official opening as well as the later relocation, while standard RAG gives only a vague year-based answer.}
\label{fig:case_study_1}
\end{figure*}

\begin{figure*}[h]
\centering
\begin{tcolorbox}[
  colback=gray!5!white,
  colframe=gray!40!black,
  width=\linewidth,
  boxrule=0.5pt,
  arc=2pt,
  outer arc=2pt,
  title=Case Study 2,
]
\paragraph{Query.} 
\textit{Where did Italy win the World Cup in 2006?}

\paragraph{Retrieved Chunks.}
\begin{verbatim}

[Chunk 1] The 2006 FIFA World Cup final was played on July 9, 2006, 
between Italy and France ... after a 1-1 draw, Italy won 5-3 on penalties ...

[Chunk 2] ... The final was held at the Olympiastadion in Berlin, Germany, 
with an attendance of 69,000 spectators ...

[Chunk 3] Italy claimed their fourth world title, having previously 
won in 1934, 1938, and 1982 ...

[Chunk 4] ... France had won their only previous World Cup in 1998, 
while Italy had lost finals in 1970 and 1994 ...

[Chunk 5] The match was part of the France-Italy football rivalry, 
and marked Italy's first world title in 24 years ...
\end{verbatim}

\paragraph{Standard RAG Output.}
\begin{quote}
Italy won the 2006 World Cup in Munich, Germany, after a penalty shootout against France. 
\end{quote}

\paragraph{Disco-RAG Output.}
\begin{quote}
Italy defeated France in the 2006 FIFA World Cup Final, played on July 9, 2006, at the Olympiastadion in Berlin, Germany. After a 1–1 draw, Italy prevailed 5–3 in the penalty shootout.
\end{quote}

\paragraph{Reference Answer.}
\begin{quote}
The Italy national football team defeated France and was declared world champions in the 2006 World Cup Final, the last match of the 2006 FIFA World Cup, at the Olympiastadion stadium in Berlin, Germany, on July 9, 2006. Italy had won three previous finals in 1934, 1938, 1982 and lost in 1970 and 1994, while France had won their only previous final in 1998.
\end{quote}

\paragraph{Remark.}
Standard RAG incorrectly states that the final took place in Munich, likely due to spurious retrieval or shallow aggregation of context. In contrast, \texttt{Disco-RAG} integrates evidence across multiple chunks, correctly identifying the Olympiastadion in Berlin as the venue and providing richer historical context.
\end{tcolorbox}
\caption{Case study comparing standard RAG and our proposed \texttt{Disco-RAG} 
on the query ``Where did Italy win the World Cup in 2006?''. Our method correctly identifies the Olympiastadion in Berlin, while standard RAG produces a factual error.}
\label{fig:case_study_2}
\end{figure*}

\begin{figure*}[h]
\centering
\begin{tcolorbox}[
  colback=gray!5!white,
  colframe=gray!40!black,
  width=\linewidth,
  boxrule=0.5pt,
  arc=2pt,
  outer arc=2pt,
  title=Case Study 3,
]
\paragraph{Query.}
\textit{Is Pluto a planet?}

\paragraph{Retrieved Chunks.}
\begin{verbatim}
[Chunk 1] Pluto was discovered in 1930 and was long regarded as the ninth planet
in the Solar System in many textbooks and popular references.

[Chunk 2] In 2006, the International Astronomical Union adopted a formal definition
of "planet" and reclassified Pluto as a dwarf planet.

[Chunk 3] The 2006 definition requires a planet to orbit the Sun, be nearly round,
and clear its orbital neighborhood, which Pluto does not satisfy.
\end{verbatim}

\paragraph{Standard RAG Output.}
\begin{quote}
Yes. Pluto is the ninth planet in the Solar System.
\end{quote}

\paragraph{Disco-RAG Output.}
\begin{quote}
Pluto was historically treated as the ninth planet, but it is currently classified as a dwarf planet. In 2006, the International Astronomical Union adopted a definition of a planet that requires clearing the orbital neighborhood, and Pluto does not meet this criterion.
\end{quote}

\paragraph{Reference Answer.}
\begin{quote}
Pluto is not classified as a planet under the current astronomical definition. It was long considered the ninth planet after its discovery in 1930, but the International Astronomical Union reclassified it as a dwarf planet in 2006 because it does not clear its orbital neighborhood.
\end{quote}

\paragraph{Remark.}
Standard RAG commits to an outdated claim from Chunk 1 and produces a categorical answer that conflicts with the current definition. In contrast, \texttt{Disco-RAG} uses the \texttt{Contrast} relation to avoid merging incompatible statements and uses the \texttt{Background} relation to ground the final answer in the relevant criterion, which reduces the risk of hallucinating a definitive but incorrect conclusion under conflicting evidence.
\end{tcolorbox}
\caption{Case study showing how discourse relations affect generation under conflicting evidence. The \texttt{Contrast} relation prevents incompatible claims from being merged, and the \texttt{Background} relation provides the criterion needed for a faithful answer.}
\label{fig:case_study_3}
\end{figure*}

\section{Prompts for Disco-RAG}
Appendix \autoref{appendix:prompt_rst}, \autoref{appendix:prompt_graph}, \autoref{appendix:prompt_plan} and \autoref{appendix:prompt_generation} present the prompts used in \texttt{Disco-RAG}.

\begin{figure*}[htbp]
\centering
\begin{tcolorbox}[
  colback=gray!5!white,    
  colframe=gray!15!black,  
  coltitle=black,          
  colbacktitle=gray!25!white, 
  fonttitle=\bfseries,           
  title=Relation Definitions for Intra-chunk RST Tree Construction,
]
\textbf{Relation Definitions:} \\
- \texttt{ELABORATION}: Satellite provides additional detail or information about the nucleus. \\
- \texttt{EXPLANATION}: Satellite explains or clarifies the nucleus content. \\
- \texttt{EVIDENCE}: Satellite provides evidence or proof for the nucleus claim. \\
- \texttt{EXAMPLE}: Satellite gives a specific example of the nucleus concept. \\
- \texttt{CONTRAST}: Satellite presents opposing or contrasting information. \\
- \texttt{COMPARISON}: Satellite compares two or more entities or concepts. \\
- \texttt{CONCESSION}: Satellite acknowledges opposing viewpoint while maintaining main claim. \\
- \texttt{ANTITHESIS}: Satellite presents directly opposite or contradictory information. \\
- \texttt{CAUSE}: Satellite describes the cause of an event or situation. \\
- \texttt{RESULT}: Satellite describes the result or consequence of an action. \\
- \texttt{CONSEQUENCE}: Satellite shows the outcome following from the nucleus. \\
- \texttt{PURPOSE}: Satellite explains the intended goal or purpose. \\
- \texttt{CONDITION}: Satellite specifies conditions under which something holds. \\
- \texttt{TEMPORAL}: Satellite indicates temporal relationship between events. \\
- \texttt{SEQUENCE}: Satellite shows sequential order of events or actions. \\
- \texttt{BACKGROUND}: Satellite provides background context or setting. \\
- \texttt{CIRCUMSTANCE}: Satellite describes circumstances surrounding an event. \\
- \texttt{SUMMARY}: Satellite summarizes or generalizes the nucleus content. \\
- \texttt{RESTATEMENT}: Satellite restates the nucleus in different words. \\
- \texttt{EVALUATION}: Satellite provides evaluation or assessment of the nucleus. \\
- \texttt{INTERPRETATION}: Satellite offers interpretation of the nucleus content. \\
- \texttt{ATTRIBUTION}: Satellite attributes information to a source. \\
- \texttt{DEFINITION}: Satellite defines a term or concept. \\
- \texttt{CLASSIFICATION}: Satellite classifies or categorizes information. \\
\end{tcolorbox}
\caption{Relation Definitions for Intra-chunk RST Tree Construction.}
\label{appendix:Relation Definitions}
\end{figure*}

\begin{figure*}[htbp]
\centering
\begin{tcolorbox}[
  colback=gray!5!white,    
  colframe=gray!15!black,  
  coltitle=black,          
  colbacktitle=gray!25!white, 
  fonttitle=\bfseries,      
  title=Relation Definitions for Inter-chunk Rhetorical Graph Construction,
]
\textbf{Relation Definitions:} \\
- \texttt{SUPPORTS}: Chunk provides support or evidence for another chunk. \\
- \texttt{CONTRADICTS}: Chunk contradicts or opposes another chunk. \\
- \texttt{ELABORATES}: Chunk elaborates on information in another chunk. \\
- \texttt{EXEMPLIFIES}: Chunk provides examples for another chunk's concepts. \\
- \texttt{CAUSES}: Chunk describes causes for events in another chunk. \\
- \texttt{RESULTS\_FROM}: Chunk describes results from another chunk's events. \\
- \texttt{ENABLES}: Chunk describes what enables another chunk's situation. \\
- \texttt{PREVENTS}: Chunk describes what prevents another chunk's situation. \\
- \texttt{PRECEDES}: Chunk describes events that precede another chunk. \\
- \texttt{FOLLOWS}: Chunk describes events that follow another chunk. \\
- \texttt{SIMULTANEOUS}: Chunk describes simultaneous events with another chunk. \\
- \texttt{BACKGROUND\_FOR}: Chunk provides background context for another chunk. \\
- \texttt{GENERALIZES}: Chunk provides general principles for another chunk's specifics. \\
- \texttt{SPECIFIES}: Chunk provides specific details for another chunk's generalizations. \\
- \texttt{COMPARES\_WITH}: Chunk compares information with another chunk. \\
- \texttt{CONTRASTS\_WITH}: Chunk contrasts information with another chunk. \\
- \texttt{SUPPLEMENTS}: Chunk supplements information in another chunk. \\
- \texttt{REPLACES}: Chunk replaces or updates information in another chunk. \\
- \texttt{MOTIVATES}: Chunk provides motivation for another chunk's content. \\
- \texttt{JUSTIFIES}: Chunk justifies claims or actions in another chunk. \\
- \texttt{UNRELATED}: Chunk has no rhetorical or semantic relation to another chunk. \\
\end{tcolorbox}
\caption{Relation Definitions for Inter-chunk Rhetorical Graph Construction.}
\label{appendix:Relation Definitions2}
\end{figure*}

\begin{figure*}[h]
\centering
\begin{tcolorbox}[width=\linewidth, 
  colback=gray!5!white,    
  colframe=gray!15!black,  
  coltitle=black,          
  colbacktitle=gray!25!white, 
  fonttitle=\bfseries,        
  title=Prompt for Intra-chunk RST Tree Construction
]
You are an expert in Rhetorical Structure Theory (RST) analysis. Your task is to analyze the given text and construct a precise RST tree.\\
\textbf{Critical instructions:} \\
1. RST tree is a hierarchical tree structure (not a graph or network). \\
2. Each internal node has exactly two children: one nucleus (core) and one satellite (support) or two nuclei at the same time. \\
3. The nucleus contains the main information; the satellite provides supporting content. \\
4. Relations describe how the satellite relates to the nucleus. \\
5. Think carefully and output ONLY ONE complete RST tree. \\
\textbf{Allowed RST relations:} \\
ELABORATION, EXPLANATION, EVIDENCE, EXAMPLE, CONTRAST, COMPARISON, CONCESSION, ANTITHESIS, CAUSE, RESULT, CONSEQUENCE, PURPOSE, CONDITION, TEMPORAL, SEQUENCE, BACKGROUND, CIRCUMSTANCE, SUMMARY, RESTATEMENT, EVALUATION, INTERPRETATION, ATTRIBUTION, DEFINITION, CLASSIFICATION \\
\textbf{Relation definitions:} \\
\{Relation Definition\} \\
\textbf{Step-by-step process:} \\
1. Segment text into meaningful elementary discourse units (EDUs). \\
2. Determine the most important EDU (this becomes the root nucleus). \\
3. For each other EDU, decide: Is it a nucleus (core) or a satellite (support)? \\
4. Assign one relation from the allowed list. \\
5. Build the binary tree bottom-up.\\
\textbf{Required output format:} \\
\texttt{EDUs:} \\
$[$1$]$ $<$first EDU$>$ \\
$[$2$]$ $<$second EDU$>$ \\
\dots \\
$[$N$]$ $<$Nth EDU$>$ \\
\texttt{RST ANALYSIS:}\\
RELATION(EDU$_{i}$, EDU$_{j}$): \{\texttt{RELATION TYPE}\} \\
\dots
\\
\texttt{TREE STRUCTURE:}
\begin{verbatim}
ROOT[1-N]
|--- NUCLEUS[X] <EDU text> (N)
|--- SATELLITE[Y] <EDU text> (S): {RELATION TYPE}
\end{verbatim}
\textbf{Validation rules:}\\
- Each EDU must be complete and meaningful. \\
- Relations must be chosen from the allowed list. \\
- Mark (N) for nucleus, (S) for satellite. \\
- Output exactly ONE complete tree. \\
\texttt{TEXT TO ANALYZE:} $\{$chunk$_{i}$$\}$
\end{tcolorbox}
\caption{Prompt for Intra-chunk RST Tree Construction. The relation definitions are provided in \autoref{appendix:Relation Definitions}.}
\label{appendix:prompt_rst}
\end{figure*}

\begin{figure*}[t]
\centering
\begin{tcolorbox}[width=\linewidth, 
  colback=gray!5!white,    
  colframe=gray!15!black,  
  coltitle=black,          
  colbacktitle=gray!25!white, 
  fonttitle=\bfseries,          
  title=Prompt for Listwise Discourse Relation Inference,
]
You are an expert in discourse analysis. Your task is to infer the rhetorical relations jointly among a list of retrieved text chunks. In each call to this prompt, you are given the entire set of chunks, and you must construct a directed rhetorical graph over all of them.

\textbf{Task objective:} \\
Given a list of chunks \texttt{CHUNK[1]}, \texttt{CHUNK[2]}, \dots, \texttt{CHUNK[K]}, determine for every ordered pair of distinct chunks whether there exists a rhetorical relation from the source chunk \texttt{CHUNK[i]} to the target chunk \texttt{CHUNK[j]}. If a relation exists, assign a directed discourse label; otherwise, mark the pair as \texttt{UNRELATED}.

\textbf{Relation direction:} \\
For each ordered pair $(i,j)$ with $i \neq j$, treat \texttt{CHUNK[i]} as the source and \texttt{CHUNK[j]} as the target. The relation type should reflect how the source chunk contributes rhetorically to the target.

\textbf{Allowed relation types:} \\
SUPPORTS, CONTRADICTS, ELABORATES, EXEMPLIFIES, CAUSES, RESULTS\_FROM, ENABLES, PREVENTS, PRECEDES, FOLLOWS, SIMULTANEOUS, BACKGROUND\_FOR, GENERALIZES, SPECIFIES, COMPARES\_WITH, CONTRASTS\_WITH, SUPPLEMENTS, REPLACES, MOTIVATES, JUSTIFIES, UNRELATED \\
\textbf{Relation definitions:} \\
\{Relation Definition\} \\
\textbf{Step-by-step process:} \\
1. Carefully read all chunks in the list and identify the main claim, fact, or event expressed in each one. \\
2. Reason about how each chunk relates to the others at the discourse level, taking into account global context across all chunks. \\
3. For every ordered pair of distinct indices $(i,j)$, decide whether \texttt{CHUNK[i]} serves a discourse function relative to \texttt{CHUNK[j]}. \\
4. If a rhetorical link exists, assign exactly one relation type from the allowed list.

\textbf{Required output format:} \\
For each ordered pair $(i,j)$ with $i \neq j$, output one line in the following format: \\
\texttt{CHUNK[i] -> CHUNK[j]: \{\texttt{RELATION\_TYPE}\}} \\
List all such lines for all ordered pairs in a consistent order (e.g., sorted by $i$ then $j$).

\textbf{Validation rules:} \\
- Use only the allowed relation types. \\
- Relation direction must be from \texttt{CHUNK[i]} to \texttt{CHUNK[j]}. \\
- Output exactly one relation type for every ordered pair with $i \neq j$. 

\texttt{TEXT TO ANALYZE}: \\
\texttt{CHUNK[1]}: [first chunk] \\
\texttt{CHUNK[2]}: [second chunk] \\
\dots \\
\texttt{CHUNK[K]}: [K-th chunk] \\
\end{tcolorbox}
\caption{
Prompt for listwise discourse relation inference. The relation definitions are provided in \autoref{appendix:Relation Definitions2}.
}
\label{appendix:prompt_graph}
\end{figure*}

\begin{figure*}[t]
\centering
\begin{tcolorbox}[width=\linewidth, 
  colback=gray!5!white,    
  colframe=gray!15!black,  
  coltitle=black,          
  colbacktitle=gray!25!white, 
  fonttitle=\bfseries,      
  title=Prompt for Discourse-Driven Planning,
]
You are an expert in discourse-aware text generation. Your task is to produce a discourse-aware plan --- a natural language paragraph that outlines how the final answer should be organized. \\
\textbf{Inputs:} \\
1. The user query. \\
2. Retrieved text chunks. \\
3. Intra-chunk RST trees, capturing local rhetorical hierarchies. \\
4. Inter-chunk rhetorical graph, modeling cross-passage discourse flow. \\
\textbf{Critical instructions:} \\
1. The plan must be written as a continuous paragraph in natural language. \\
2. The plan should describe the intended organization of the final answer. \\
3. The plan must be dynamically adapted to the given user query and evidence. \\
4. Avoid reproducing the content of the chunks; only outline how they will be used. \\
5. Output exactly one complete rhetorical plan. \\
\textbf{Required output format:} \\
\texttt{PLAN:} $<$one paragraph in natural language that describes the planned organization of the answer$>$ \\
\texttt{TEXT TO ANALYZE}: $\{$query, chunks, RST trees, rhetorical graph$\}$
\end{tcolorbox}
\caption{Prompt for Discourse-Driven Planning.}
\label{appendix:prompt_plan}
\end{figure*}

\begin{figure*}[t]
\centering
\begin{tcolorbox}[width=\linewidth, 
  colback=gray!5!white,    
  colframe=gray!15!black,  
  coltitle=black,          
  colbacktitle=gray!25!white, 
  fonttitle=\bfseries,     
  title=Prompt for Full Context Generation,
]
You are an expert in question answering and text generation. Your task is to answer the user query using the provided full document as context. \\
\textbf{Inputs:} \\
1. The user query. \\
2. The full document. \\
\textbf{Critical instructions:} \\
1. The answer must directly address the user query. \\
2. Use the full document as the only source of factual claims. \\
3. If the document does not support a claim, do not add it. \\
4. Write a coherent answer without copying long spans verbatim from the document. \\
\textbf{Required output format:} \\
\texttt{ANSWER} $<$one paragraph or multiple paragraphs in natural language$>$ \\
\texttt{TEXT TO ANALYZE} $\{$query, document$\}$
\end{tcolorbox}
\caption{Prompt for full context generation used in our baseline.}
\label{appendix:prompt_full_context}
\end{figure*}

\begin{figure*}[t]
\centering
\begin{tcolorbox}[width=\linewidth, 
  colback=gray!5!white,    
  colframe=gray!15!black,  
  coltitle=black,          
  colbacktitle=gray!25!white, 
  fonttitle=\bfseries,     
  title=Prompt for Standard RAG,
]
You are an expert in retrieval-augmented generation. Your task is to answer the user query using only the retrieved text chunks as evidence. \\
\textbf{Inputs:} \\
1. The user query. \\
2. Retrieved text chunks. \\
\textbf{Critical instructions:} \\
1. The answer must directly address the user query. \\
2. Use the retrieved chunks as the only source of factual claims. \\
3. If the retrieved chunks do not support a claim, do not add it. \\
4. Write a coherent answer without copying long spans verbatim from the chunks. \\
\textbf{Required output format:} \\
\texttt{ANSWER} $<$one paragraph or multiple paragraphs in natural language$>$ \\
\texttt{TEXT TO ANALYZE} $\{$query, chunks$\}$
\end{tcolorbox}
\caption{Prompt for standard RAG used in our baseline.}
\label{appendix:prompt_standard_rag}
\end{figure*}

\begin{figure*}[t]
\centering
\begin{tcolorbox}[width=\linewidth,
  colback=gray!5!white,
  colframe=gray!15!black,
  coltitle=black,
  colbacktitle=gray!25!white,
  fonttitle=\bfseries,
  title=Prompt for Retrieve-and-Plan Baseline
]
You are an expert in retrieval-augmented generation. Your task is to answer the user query by first writing a short plan and then writing the final answer using only the retrieved text chunks as evidence. \\
\textbf{Inputs:} \\
1. The user query. \\
2. Retrieved text chunks. \\
\textbf{Critical instructions:} \\
1. Write the plan as a single continuous paragraph that outlines the structure of the answer. \\
2. The plan must be grounded in what is supported by the retrieved chunks. \\
3. The answer must directly address the user query and use the retrieved chunks as the only source of factual claims. \\
4. If the retrieved chunks do not support a claim, do not add it. \\
5. Write a coherent answer without copying long spans verbatim from the chunks. \\
\textbf{Required output format:} \\
\texttt{PLAN} $<$one paragraph plan$>$ \\
\texttt{ANSWER} $<$one paragraph or multiple paragraphs in natural language$>$ \\
\texttt{TEXT TO ANALYZE} $\{$query, chunks$\}$
\end{tcolorbox}
\caption{Prompt for the retrieve-and-plan baseline used in our ablation study.}
\label{appendix:prompt_retrieve_and_plan}
\end{figure*}

\begin{figure*}[t]
\centering
\begin{tcolorbox}[width=\linewidth,
  colback=gray!5!white,
  colframe=gray!15!black,
  coltitle=black,
  colbacktitle=gray!25!white,
  fonttitle=\bfseries,
  title=Prompt for Plan-and-Retrieve Baseline
]
You are an expert in retrieval-augmented generation. Your task is to support a plan-guided retrieval procedure and then answer the user query. \\
\textbf{Stage 1} \\
Given only the user query, write a short plan and a retrieval hint that summarizes what evidence should be retrieved. \\
\textbf{Stage 2} \\
After retrieving all text chunks using the retrieval hint, write the final answer using only the retrieved text chunks as evidence. \\
\textbf{Inputs:} \\
1. The user query. \\
2. Retrieved text chunks returned after plan-guided retrieval. \\
\textbf{Critical instructions:} \\
1. Write the plan as a single continuous paragraph that outlines the structure of the answer. \\
2. The retrieval hint must be a list of retrieval queries that help retrieve evidence aligned with the plan. \\
3. The answer must directly address the user query and use the retrieved chunks as the only source of factual claims. \\
4. If the retrieved chunks do not support a claim, do not add it. \\
5. Write a coherent answer without copying long spans verbatim from the chunks. \\
\textbf{Required output format:} \\
\texttt{PLAN} $<$one paragraph plan$>$ \\
\texttt{RETRIEVAL HINT} $<$a list of retrieval queries$>$ \\
\texttt{ANSWER} $<$one paragraph or multiple paragraphs in natural language$>$ \\
\texttt{TEXT TO ANALYZE} $\{$query, chunks$\}$
\end{tcolorbox}
\caption{Prompt for the plan-and-retrieve baseline used in our ablation study.}
\label{appendix:prompt_plan_and_retrieve}
\end{figure*}

\begin{figure*}[t]
\centering
\begin{tcolorbox}[width=\linewidth,
  colback=gray!5!white,
  colframe=gray!15!black,
  coltitle=black,
  colbacktitle=gray!25!white,
  fonttitle=\bfseries,
  title=Prompt for Shallow Discourse Marker Inference
]
You are an expert in discourse analysis. Your task is to infer explicit discourse markers jointly among a list of retrieved text chunks. In each call to this prompt, you are given the entire set of chunks, and you must output a marker decision for every ordered pair of distinct chunks. \\
\textbf{Task objective:} \\
Given a list of chunks \texttt{CHUNK[1]}, \texttt{CHUNK[2]}, \dots, \texttt{CHUNK[K]}, determine for every ordered pair $(i,j)$ with $i \neq j$ whether there exists an explicit discourse marker from a marker list that indicates a rhetorical connection from \texttt{CHUNK[i]} to \texttt{CHUNK[j]}. If no marker is supported, output \texttt{NONE}. \\
\textbf{Discourse marker list:} \\
\textit{however}, \textit{but}, \textit{although}, \textit{in contrast}, \textit{therefore}, \textit{because}, \textit{as a result}, \textit{meanwhile}, \textit{moreover}, \textit{furthermore}, \textit{for example}, \textit{for instance}, \textit{in addition} \\
\textbf{Critical instructions:} \\
1. For each ordered pair $(i,j)$ with $i \neq j$, treat \texttt{CHUNK[i]} as the source and \texttt{CHUNK[j]} as the target. \\
2. Consider only explicit connectives that are supported by the two chunks. Do not infer implicit relations. \\
3. Output exactly one marker from the marker list if a marker is applicable; otherwise, output \texttt{NONE}. \\
4. Output a decision for every ordered pair with $i \neq j$. \\
\textbf{Required output format:} \\
For each ordered pair $(i,j)$ with $i \neq j$, output one line in the following format: \\
\texttt{CHUNK[i] -> CHUNK[j]: \{\texttt{MARKER}\}} \\
\texttt{TEXT TO ANALYZE}: \\
\texttt{CHUNK[1]}: [first chunk] \\
\texttt{CHUNK[2]}: [second chunk] \\
\dots \\
\texttt{CHUNK[K]}: [K-th chunk] \\
\end{tcolorbox}
\caption{Prompt for discourse marker inference used in the shallow discourse marker baseline.}
\label{appendix:prompt_shallow_marker}
\end{figure*}

\begin{figure*}[t]
\centering
\begin{tcolorbox}[width=\linewidth, 
  colback=gray!5!white,    
  colframe=gray!15!black,  
  coltitle=black,          
  colbacktitle=gray!25!white, 
  fonttitle=\bfseries,     
  title=Prompt for Discourse-Guided RAG,
]
You are an expert in retrieval-augmented generation with discourse knowledge. 
Your task is to generate a coherent and faithful answer by leveraging the following inputs: \\
\textbf{Inputs:} \\
1. The user query. \\
2. Retrieved text chunks. \\
3. Intra-chunk RST trees, capturing local rhetorical hierarchies. \\
4. Inter-chunk rhetorical graph, modeling cross-passage discourse flow. \\
5. A discourse-aware plan that outlines the intended argumentative organization. \\
\textbf{Critical instructions:} \\
1. The answer must directly address the user query. \\
2. Integrate evidence from multiple chunks, guided by their RST trees and rhetorical graph. \\
3. Follow the discourse-aware plan for structuring the answer. \\
4. Maintain factual accuracy, logical coherence, and rhetorical clarity. \\
5. Output a continuous answer in natural language. \\
\textbf{Required output format:} \\
\texttt{ANSWER:} $<$a single coherent paragraph or multi-paragraph answer grounded in discourse structures$>$ \\
\textbf{Validation requirements:} \\
- The answer must be faithful to the retrieved content. \\
- The answer must be logically organized and reflect discourse-level coherence. \\
- Avoid verbatim repetition of chunks; instead, synthesize and integrate them. \\
- Output exactly one complete answer. \\
\texttt{TEXT TO ANALYZE}: $\{$query, chunks, RST trees, rhetorical graph, discourse-aware plan$\}$ \\
\end{tcolorbox}
\caption{Prompt for Discourse-Guided RAG.}
\label{appendix:prompt_generation}
\end{figure*}

\section{Human Evaluation Guidelines}
\label{human_evaluation_guideline}
\begin{figure*}[h]
\centering
\begin{tcolorbox}[width=\linewidth, 
  colback=gray!5!white,    
  colframe=gray!15!black,  
  coltitle=black,          
  colbacktitle=gray!25!white, 
  fonttitle=\bfseries,
  title=Human Evaluation Guidelines]
  
\paragraph{Prerequisites:} Eligibility for this evaluation requires simultaneous fulfillment of two conditions: being a Master's or Ph.D. student in Computer Science or a closely related field, and demonstrating advanced proficiency in English sufficient to read and assess scientific news articles. Participants are compensated at the standard hourly rate and are asked to confirm that they meet these criteria before taking part in the task.

\paragraph{Instructions:} For each selected sample, annotators are given the source document together with four anonymized summaries, and the system identities are hidden, and the order is randomized for every instance. Raters are instructed to first read the source document carefully and then evaluate each summary independently using a three-point Likert scale along four criteria: \textit{Relevance}, \textit{Simplicity}, \textit{Conciseness}, and \textit{Faithfulness}.

\paragraph{Evaluation Criteria:} Below, we provide a detailed explanation of the four criteria used in our human evaluation. Raters are asked to consider each criterion separately and to base their scores only on the information that is explicitly supported by the source document.

\begin{itemize}[leftmargin=8pt,itemsep=1pt,topsep=1pt,parsep=1pt]
    \item \textbf{Relevance} This criterion assesses how well the summary covers the main topics, events, and findings discussed in the source document. A highly relevant summary focuses on central points and avoids spending space on marginal or tangential details.
    \item \textbf{Simplicity} This criterion measures how easy the summary is to read and understand. A simple summary uses clear and precise language, maintains a coherent structure, and avoids unnecessary jargon or convoluted phrasing that could hinder comprehension.
    \item \textbf{Conciseness} This criterion evaluates whether the summary is compact while still conveying the essential content. A concise summary avoids repetition and digression, omits minor details that are not needed for understanding, and does not exceed the length required to communicate the core message.
    \item \textbf{Faithfulness} This criterion judges whether the summary is supported by the source document and free of hallucinations. A faithful summary does not introduce claims that contradict the source, does not exaggerate or overgeneralize findings, and does not omit critical qualifications that change the meaning of the original text.
\end{itemize}

\paragraph{Rating System:} For each criterion, raters assign an integer score from 1 to 3, where 1 indicates low quality, 2 indicates acceptable quality, and 3 indicates high quality. Scores should be given solely based on the source document and the summary, without using AI tools to assist in judgment. Annotators may consult trusted external resources, such as textbooks or scientific encyclopedias, only when they need to clarify terminology.

\end{tcolorbox}
\caption{Guidelines presented to human raters for the SciNews dataset evaluation.}
\end{figure*}

\end{document}